\documentclass[twoside]{article}

\usepackage[accepted]{aistats2019}
\usepackage{hyperref}       % hyperlinks
\usepackage{url}            % simple URL typesetting
\usepackage{booktabs}       % professional-quality tables
\usepackage{amsfonts}       % blackboard math symbols
\usepackage{nicefrac}       % compact symbols for 1/2, etc.
\usepackage{microtype}      % microtypography
\usepackage[pdftex]{graphicx}
\usepackage{amsmath}
\usepackage[round]{natbib}

\begin{document}

\runningtitle{Scalable Multi-Task Gaussian Process Tensor Regression}

\runningauthor{Kia, Beckmann, Marquand}

\twocolumn[

\aistatstitle{Scalable Multi-Task Gaussian Process Tensor Regression for Normative Modeling of Structured Variation in Neuroimaging Data}

\aistatsauthor{ Seyed Mostafa Kia \And Christian F. Beckmann \And  Andre F. Marquand}

\aistatsaddress{ RadboudUMC, \\ Donders Institute, \\ Nijmegen, The Netherlands \And  RadboudUMC, \\ Donders Institute, \\ University of Oxford \And RadboudUMC, \\ Donders Institute \\ King's College London } 
]
\begin{abstract}
Most brain disorders are very heterogeneous in terms of their underlying biology and developing analysis methods to model such heterogeneity is a major challenge. A promising approach is to use probabilistic regression methods to estimate normative models of brain measures then use these to map variation across individuals. To fully capture individual differences and detect disorders in individual subjects it is crucial to statistically model patterns of correlation across different brain regions and individuals. However, this is very challenging for neuroimaging data because of high dimensionality and highly structured correlations across multiple axes. Here, we propose \emph{tensor Gaussian predictive process} (TGPP) as a general and flexible Bayesian mixed-effects modeling framework. In TGPP, we develop \emph{multi-task Gaussian process tensor regression} (MT-GPTR) to simultaneously model the structured random effects and structured noise. We use Kronecker algebra and a low-rank approximation to efficiently scale MT-GPTR to the whole brain. On a publicly available clinical fMRI dataset and in a novelty detection scenario, we show that our computationally affordable multivariate normative modeling approach substantially improves the detection rate over a baseline mass-univariate normative model and an off-the-shelf supervised alternative.
\end{abstract}

\section{Introduction} 
\label{sec:introduction}
% Why neuroimaging
Neuroimaging techniques provide detailed measures of brain structure and function which can serve as candidate biomarkers for brain disorders. However, these data present substantial challenges including: i) variability along multiple axes including across different individuals, brain locations, and cognitive systems~\citep{gratton2018functional}; ii) high dimensionality, where a large number of measurements (order of $10^{5-6}$) are acquired from multiple subjects (order of $10^{2-3}$); iii) strong correlations within and across data axes. There is a pressing need to develop methods that can model such complex covariance structures and that scale reasonably with increasing computational demands. 

Recently, there has been great interest in applying machine learning methods to quantify biological measures (biomarkers) to assist medical decision making; for example assisting diagnosis or predicting treatment outcome in the spirit of \emph{precision medicine}~\citep{Mirnezami2012preparing}. In psychiatry, this is very challenging because the diagnosis is typically based on clinical symptoms and the underlying biology is highly heterogeneous~\citep{kapur2012has}. For example, subjects with the same diagnosis may have different underlying biological signatures. Most research ignores such heterogeneity and instead regards groups as distinct entities~\citep{foulkes2018studying}, \emph{e.g.}, in a case-control approach where subjects are either ``patients" or ``controls". Supervised machine learning methods applied to neuroimaging data have been widely used for this but their accuracy is fundamentally limited by the heterogeneity within each disorder~\citep{wolfers2015estimating}, therefore, there is an urgent need to go beyond case-control settings. Normative modeling~\citep{marquand2016understanding} is a promising approach for this that aims to characterize variation across a healthy cohort before making predictions so that subjects that deviate from the resulting \emph{normative} model can be detected as outliers (\emph{i.e.}, in a novelty detection setting) and the pattern underlying the deviation can be mapped to understand the biological underpinnings. 

% Tensor regression, Gaussian Process, anomaly detection
Bayesian inference is an important component of normative modeling as it provides coherent estimates of predictive confidence. The original normative modeling approach proposed in~\citep{marquand2016understanding} uses Gaussian process regression~\citep{williams1996gaussian} (GPR) to independently regress neuroimaging measures, such as a single voxel value, on clinical covariates. Therefore, this is done in a mass-univariate fashion ignoring correlations between sampled brain locations. Since the biological signature of different disorders may be encoded via correlations between variables, this approach is sub-optimal. This problem can be mitigated by using multi-task GPR (MT-GPR)~\citep{bonilla2008multi} to jointly predict multiple brain measurements. However, applying MT-GPR on neuroimaging data is very computationally demanding because of the need to invert large covariance matrices across space, subjects or both~\citep{bowman2008bayesian}. Various approaches have been proposed in the literature to improve the computational efficiency of MT-GPR using approximations~\citep{alvarez2009sparse,alvarez2010efficient,alvarez2011computationally} or utilizing properties of Kronecker product~\citep{stegle2011efficient,rakitsch2013all}. However, MT-GPR remains computationally intractable in processing neuroimaging data at the whole-brain level.

% Our contributions
The aim of this paper is to find a principled solution for multivariate normative modeling on multi-way neuroimaging data. In this direction, we make four contributions: i) considering the tensor structure of neuroimaging data and assuming a tensor-variate normal distribution on the random-effect and noise, we propose \emph{tensor Gaussian predictive process} (TGPP) as a general and versatile Bayesian mixed-effects modeling framework. This can be seen as a generalization of previous approaches such as the spatial Gaussian predictive process (SGPP) framework~\citep{hyun2014SGPP,hyun2016STGP}. Thus, it can easily be extended to handle additional sources of variation (\emph{e.g.}, across timepoints or data modalities). This framework allows us to jointly predict multiple output dimensions, accounting for correlations within and across dimensions and potentially heteroscedastic noise structures. ii) Within the TGPP framework, we propose \emph{multi-task Gaussian process tensor regression} (MT-GPTR) approach to simultaneously learn the covariance structure of the random-effect and noise. MT-GPTR generalizes application of previous approaches that use a Kronecker product covariance structure, \emph{e.g.}, ``GP-Kronsum"~\citep{rakitsch2013all}, to multi-way tensor structured data with arbitrary dimensions. iii) Using low-rank approximation of the high-dimensional task covariance matrix via tensor factorization techniques~\citep{morup2011applications} and further exploiting algebraic properties of the Kronecker product~\citep{loan2000ubiquitous}, we develop scalable MT-GPTR (sMT-GPTR) which scales up to simultaneously predicting hundreds of thousands of tasks (\emph{i.e.}, the whole brain) using reasonable time and space resources. iv) Finally, we present an application of sMT-GPTR to normative modeling of structured variation in neuroimaging data. To this end, we apply it to a publicly available clinical fMRI dataset~\citep{poldrack2016phenome} in order to jointly predict task-related fMRI brain activity from a set of clinical covariates in a mixed-effects modeling paradigm~\citep{friston1999multisubject}. Our experimental results show that sMT-GPTR is effective and feasible in modeling variation across both space and subjects in a healthy human cohort using whole-brain neuroimaging data. In addition, in an unsupervised novelty detection scenario, the proposed method more accurately identifies psychiatric patients from healthy individuals compared to mass-univariate normative modeling and a supervised support vector machine classifier. In other words, our approach trained only on healthy participants performs better at detecting abnormal samples than a supervised approach that has full access to the diagnostic labels.
\section{Methods} \label{sec:methods}
\subsection{Notation} 
\label{subsec:notation}
In this text, we use respectively calligraphic capital letters, $\mathcal{A}$, boldface capital letters, $\mathbf{A}$, and capital letters, $A$, to denote tensors, matrices, and scalar numbers. We denote the vertical vector which results from collapsing a matrix $\mathbf{A}$ or tensor $\mathcal{A}$ with $vec(\mathbf{A})$ or $vec(\mathcal{A})$, respectively. We denote an identity matrix by $\mathbf{I}$; and the determinant, diagonal elements, and the trace of matrix $\mathbf{A}$ with $\left | \mathbf{A} \right |$, $diag(\mathbf{A})$, and $Tr[\mathbf{A}]$, respectively. We use $\otimes$, $\odot$, and $\times_n$ to respectively denote Kronecker, element-wise, and $n$-mode tensor products. The $i$-mode matricized version of a tensor $\mathcal{A} \in \mathbb{R}^{I_1 \times \dots \times I_D}$ is shown as $\mathbf{A}_{(i)} \in \mathbb{R}^{I_i \times I_1 \dots I_{i-1} I_{i+1} \dots I_D}$. We use concise notation $\mathbf{A}_i \mid_{i=1}^D$ and $\bigotimes_{i=1}^D \mathbf{A}_i$ for $\mathbf{A}_1, \dots, \mathbf{A}_D$ and $\mathbf{A}_1 \otimes \dots \otimes \mathbf{A}_D$, respectively. We use $\mathbf{A}[i,j]$, $\mathbf{A}[:,i]$, and $\mathbf{A}[i,:]$ to refer to a certain element, row, or column vector in a matrix $\mathbf{A}$ (similar for a tensor $\mathcal{A}$). 

\subsection{Tensor Gaussian Predictive Process for Modeling Neuroimaging Data} 
\label{subsec:TGPP}
Consider a neuroimaging study with $N$ subjects and let $\mathbf{X} \in \mathbb{R}^{N \times F}$ to denote the design matrix of $F$ covariates of interest for $N$ subjects (\emph{e.g.}, demographic, cognitive, or clinical variables). Let $\mathcal{Y} \in \mathbb{R}^{N \times T_1 \times \dots \times T_D}$ to represent a $(D+1)$-order tensor of multivariate neuroimaging data for corresponding $N$ subjects. In this text, we refer to $D$ as the number of dimensions of multi-way neuroimaging data. For example in the case of volumetric structural MRI, we have $D=3$ where each dimension refers to $x$, $y$, and $z$ axis, hence $\mathcal{Y}$ is a 4-order tensor with $T_1$, $T_2$, and $T_3$ voxels in corresponding data dimensions. From the theoretical perspective, we put no restriction on the order of $\mathcal{Y}$ and it could take any value between 2 to an arbitrary natural number. This makes the presented methodology very flexible for different neuroimaging modalities (\emph{e.g.}, structural/functional MRI) and study designs (\emph{e.g.}, longitudinal studies, multiple contrast data). Extending Gaussian predictive process models~\citep{hyun2014SGPP,hyun2016STGP} and as a generalization of the general linear model (GLM) to multi-way data structures, we define the tensor Gaussian predictive process (TGPP) as follows:
\begin{eqnarray} \label{eq:TGPP}
\mathcal{Y} = \mathbf{X} \times_1 \mathcal{A} + \mathcal{Z} + \mathcal{E} \quad ,
\end{eqnarray}

\noindent where $\mathcal{A} \in \mathbb{R}^{F \times T_1 \times \dots \times T_D}$ is a $(D+1)$-order tensor that contains regression coefficients estimated by solving the following linear equations (for example using ordinary least squares regression):
\small
\begin{align*} \label{eq:fixed_effect}
\begin{split}
\hat{\mathcal{Y}}[:, i,\dots , j] = \mathbf{X} \mathcal{A}[:, i,\dots , j], \quad & for \quad i=1,\dots, T_1; \dots ; \\ & for \quad j=1,\dots, T_D .
\end{split}
\end{align*}
\normalsize
Here, $\mathcal{A}$ represents the \emph{fixed-effect} across subjects. On the other hand, $\mathcal{Z} \in \mathbb{R}^{N \times T_1 \times \dots \times T_D}$ represents the \emph{random-effect} that characterizes the joint variations from the fixed-effect across different dimensions of neuroimaging data in $\mathcal{Y}$ (\emph{e.g.}, across different individuals, spatio-temporal measures, or modalities). Finally, $\mathcal{E} \in \mathbb{R}^{N \times T_1 \times \dots \times T_D}$ is multivariate structured noise. In the TGPP framework, without loss of generality we assume a zero-mean tensor-variate normal distribution, as a generalization of the matrix normal distribution, for $\mathcal{Z}$ and $\mathcal{E}$:
\small
\begin{subequations} \label{eq:TND}
\begin{align}
\begin{split}
p(\mathcal{Z} & \mid \mathbf{D}_i \mid_{i=1}^D, \mathbf{R}) = \mathcal{TN} (\mathbf{0}, \bigotimes_{i=1}^D \mathbf{D}_i \otimes \mathbf{R})= \\
& \frac{\exp(-\frac{1}{2}vec(\mathcal{Z})^\top [\bigotimes_{i=1}^D \mathbf{D}_i \otimes \mathbf{R}]^{-1}vec(\mathcal{Z}))}{\sqrt{(2\pi)^{NT} \left | \bigotimes_{i=1}^D \mathbf{D}_i \right |^{N} \left | \mathbf{R} \right |^T}}  ,
\end{split}
\\
\begin{split}
p(\mathcal{E} & \mid \mathbf{\Xi}_i \mid_{i=1}^D, \mathbf{\Omega}) = \mathcal{TN} (\mathbf{0}, \bigotimes_{i=1}^D \mathbf{\Xi}_i \otimes \mathbf{\Omega})= \\
& \frac{\exp(-\frac{1}{2}vec(\mathcal{E})^\top [\bigotimes_{i=1}^D \mathbf{\Xi}_i \otimes \mathbf{\Omega}]^{-1}vec(\mathcal{E}))}{\sqrt{(2\pi)^{NT} \left | \bigotimes_{i=1}^D \mathbf{\Xi}_i \right |^{N} \left | \mathbf{\Omega} \right |^T}} ,
\end{split}
\end{align}
\end{subequations} 
\normalsize
\noindent where $T = \prod_{i=1}^{D}T_i$, and $\mathbf{R},\mathbf{\Omega} \in \mathbb{R}^{N \times N}$ are respectively covariance matrices of $\mathcal{Z}$ and $\mathcal{E}$ across subjects; $\mathbf{D}_i, \mathbf{\Xi}_i \in \mathbb{R}^{T_i \times T_i}$ represent the covariance matrices of random-effect and noise terms across $i$th dimension of data, \emph{i.e.}, $i$-mode covariance matrices of $\mathcal{Z}$ and $\mathcal{E}$. Based on this assumption on the distribution of $\mathcal{Z}$ and $\mathcal{E}$, we generalize sum of Kronecker products covariance structure (GP-Kronsum) approach~\citep{rakitsch2013all} to the multi-task Gaussian process tensor regression (MT-GPTR) to jointly estimate parameters of $\mathbf{R},\mathbf{\Omega},\mathbf{D}_i,$ and $\mathbf{\Xi}_i$ in a multi-way representation of neuroimaging data:
\small
\begin{align} \label{eq:kronsum}
\begin{split}
& p(vec(\mathcal{Z}+\mathcal{E}) \mid \mathbf{X}, \mathbf{D}_i\mid_{i=1}^D, \mathbf{R},\mathbf{\Xi}_i\mid_{i=1}^D,\mathbf{\Omega})= \\ 
& \mathcal{GP} (vec(\mathcal{Y}-\hat{\mathcal{Y}}) \mid \mathbf{0},\bigotimes_{i=1}^D \mathbf{D}_i \otimes \mathbf{R} + \bigotimes_{i=1}^D \mathbf{\Xi}_i \otimes \mathbf{\Omega}).
\end{split}
\end{align}
\normalsize
Here $\mathbf{R}$ and $\mathbf{\Omega}$ are defined in the input space $\mathbf{X}$ in a multi-task setting~\citep{bonilla2008multi}. Considering the inherent high dimensionality of neuroimaging data, computing the inverse covariance matrix in Eq.~\ref{eq:kronsum} is computationally expensive, thus there is a pressing need to reduce the time and space complexities of MT-GPTR. In the following, we combine the tensor factorization technique with elegant properties of Kronecker product~\citep{loan2000ubiquitous} in order to extend the application of MT-GPTR to large output spaces. 

\subsection{Scalable Multi-Task Gaussian Process Tensor Regression (sMT-GPTR)} 
\label{subsec:MT-GPTR}
Let $\Phi: \mathcal{Y} - \hat{\mathcal{Y}} \to \mathcal{Z'}$ be an orthogonal linear transformation that transforms $\mathcal{Z} + \mathcal{E}$ to a reduced latent space $\mathcal{Z'} \in \mathbb{R}^{N \times P_1 \times \dots \times P_D}$, where $P_i < T_i$. A tensor factorization technique~\citep{kolda2009tensor} can be used for this transformation wherein $\mathcal{Z} + \mathcal{E} \approx \hat{\mathcal{Z}} = \mathcal{Z'} \times_2 \mathbf{B}_1 \times_3 \dots \times_{D+1} \mathbf{B}_D$. Here, columns of $\mathbf{B}_i \in \mathbb{R}^{T_i \times P_i}$ represent a set of $P_i$ orthogonal basis functions across the $i$th dimension of data. Assuming a zero-mean tensor-variate normal distribution for $\mathcal{Z'}$, we have (see supplement for the derivation):
\small
\begin{align} \label{eq:TND_Zprime}
\begin{split}
p(\mathcal{Z'}  & \mid \mathbf{C}_i \mid_{i=1}^D, \mathbf{R}) = \mathcal{TN}(\mathbf{0}, \bigotimes_{i=1}^D \mathbf{C}_i \otimes \mathbf{R}) = \\
& \frac{\exp(-\frac{1}{2}Tr[\bigotimes_{i=1}^D \mathbf{B}_i \mathbf{C}_i^{-1}\mathbf{B}_i^\top \hat{\mathbf{Z}}_{(1)}^\top \mathbf{R}^{-1} \hat{\mathbf{Z}}_{(1)}])}{\sqrt{(2\pi)^{N\prod_{i=1}^{D}P_i} \left | \bigotimes_{i=1}^D \mathbf{C}_i \right |^{N} \left | \mathbf{R} \right |^{\prod_{i=1}^{D}P_i}}} ,
\end{split}
\end{align} 
\normalsize
\noindent where $\mathbf{C}_i \in \mathbb{R}^{P_i \times P_i}$ is the $i$-mode covariance matrix in the reduced latent space. Then, we have $p(\mathcal{Z'} \mid \mathbf{C}_i \mid_{i=1}^D, \mathbf{R})=p(\hat{\mathcal{Z}} \mid \mathbf{B}_i\mathbf{C}_i\mathbf{B}_i^\top \mid_{i=1}^D, \mathbf{R})$. Assuming $\hat{\mathcal{Z}}$ to explain the majority of the variance in the random-effect, we use the numerator in Eq.~\ref{eq:TND_Zprime} as an approximation for the numerator in Eq.~\ref{eq:TND}a, thus:
\begin{eqnarray} \label{eq:TND_Z}
p(\mathcal{Z}  \mid \mathbf{D}_i \mid_{i=1}^D, \mathbf{R}) \approx p(\hat{\mathcal{Z}} \mid \mathbf{B}_i\mathbf{C}_i\mathbf{B}_i^\top \mid_{i=1}^D, \mathbf{R}) \quad , 
\end{eqnarray}

\noindent where $\mathbf{D}_i$ is approximated by $\mathbf{B}_i\mathbf{C}_i\mathbf{B}_i^\top$. Analogously, using $\mathcal{Y} - \hat{\mathcal{Y}} - \hat{\mathcal{Z}}$ as a proxy for $\mathcal{E}$ and setting $\mathcal{Y} - \hat{\mathcal{Y}} - \hat{\mathcal{Z}} \approx \hat{\mathcal{E}} = \mathcal{E'} \times_2 \mathbf{\Lambda}_1 \times_3 \dots \times_{D+1} \mathbf{\Lambda}_D$ , for $\mathbf{\Lambda}_i \in \mathbb{R}^{T_i \times Q_i}$, and assuming a zero-mean tensor-variate normal distribution on $\mathcal{E'}$ we have:
\begin{eqnarray} \label{eq:TND_E}
p(\mathcal{E}  \mid \mathbf{\Xi}_i \mid_{i=1}^D, \mathbf{\Omega}) \approx p(\hat{\mathcal{E}} \mid \mathbf{\Lambda}_i\mathbf{\Sigma}_i\mathbf{\Lambda}_i^\top \mid_{i=1}^D, \mathbf{\Omega}) \quad .
\end{eqnarray}

Based on Eq.~\ref{eq:TND_Z} and Eq.~\ref{eq:TND_E}, our scalable multi-task Gaussian process tensor regression (sMT-GPTR) model can be derived in the latent space by rewriting Eq.~\ref{eq:kronsum} using approximated covariance matrices:
\scriptsize
\begin{align} \label{eq:sMT-GPTR}
\begin{split}
& p(vec(\mathcal{Z}+\mathcal{E}) \mid \mathbf{X}, \mathbf{D}_i\mid_{i=1}^D, \mathbf{R}, \mathbf{\Xi}_i\mid_{i=1}^D \mathbf{\Omega}) \approx  \\
& p(vec(\hat{\mathcal{Z}}+\hat{\mathcal{E}}) \mid \mathbf{X}, \mathbf{C}_i\mid_{i=1}^D,\mathbf{B}_i\mid_{i=1}^D, \mathbf{R}, \mathbf{\Sigma}_i\mid_{i=1}^D, \mathbf{\Lambda}_i\mid_{i=1}^D, \mathbf{\Omega}) = \\ 
& \mathcal{GP} (vec(\mathcal{Y}-\hat{\mathcal{Y}}) \mid \mathbf{0},\bigotimes_{i=1}^D \mathbf{B}_i\mathbf{C}_i\mathbf{B}_i^\top \otimes \mathbf{R}  + \bigotimes_{i=1}^D \mathbf{\Lambda}_i\mathbf{\Sigma}_i\mathbf{\Lambda}_i^\top \otimes \mathbf{\Omega}).
\end{split}
\end{align}
\normalsize
\subsubsection{Predictive Distribution} 
\label{subsubsec:prediction}
Following the standard GPR framework~\citep{williams1996gaussian}, the mean and variance of the predictive distribution of sMT-GPTR in Eq.~\ref{eq:sMT-GPTR} on $N^*$ test samples, \emph{i.e.}, $p(vec(\mathcal{Y}^*) - vec(\hat{\mathcal{Y}}^*) \mid vec(\mathcal{M}^*), \mathbf{V}^*)$, in which $\mathbf{V}^*\in \mathbb{R}^{N^*T \times N^*T}$, can be computed as follows:
\scriptsize
\begin{subequations} \label{eq:predictive_distribution}
\begin{align}
\begin{split}
vec(\mathcal{M}^*) = ( & \bigotimes_{i=1}^D \mathbf{B}_i\mathbf{C}_i\mathbf{B}_i^\top \otimes \mathbf{R}^*)(\bigotimes_{i=1}^D \mathbf{B}_i\mathbf{C}_i\mathbf{B}_i^\top \otimes \mathbf{R} + \\ 
& \bigotimes_{i=1}^D \mathbf{\Lambda}_i\mathbf{\Sigma}_i\mathbf{\Lambda}_i^\top \otimes \mathbf{\Omega})^{-1} vec(\mathcal{Y}),
\end{split}
\\
\begin{split}
\mathbf{V}^* = & (\bigotimes_{i=1}^D \mathbf{B}_i\mathbf{C}_i\mathbf{B}_i^\top \otimes \mathbf{R}^{**})-(\bigotimes_{i=1}^D \mathbf{B}_i\mathbf{C}_i\mathbf{B}_i^\top \otimes \mathbf{R}^*) \\ 
& (\bigotimes_{i=1}^D \mathbf{B}_i\mathbf{C}_i\mathbf{B}_i^\top \otimes \mathbf{R} + \bigotimes_{i=1}^D \mathbf{\Lambda}_i\mathbf{\Sigma}_i\mathbf{\Lambda}_i^\top \otimes \mathbf{\Omega})^{-1} \\
& (\bigotimes_{i=1}^D \mathbf{B}_i\mathbf{C}_i\mathbf{B}_i^\top \otimes \mathbf{R}^{* \top}),
\end{split}
\end{align}
\end{subequations}
\normalsize
\noindent where $\mathbf{R}^{**} \in \mathbb{R}^{N^* \times N^*}$ is the covariance matrix of test samples, and $\mathbf{R}^* \in \mathbb{R}^{N^* \times N}$ is the cross-covariance matrix between the test and training samples. Importantly, sMT-GPTR enables us to estimate separate structured components for \emph{epistemic} and \emph{aleatoric} uncertainties~\citep{kendall2017uncertainties} in the output space. These respectively quantify modeling uncertainty that can be reduced given more data (\emph{e.g.}, parameter uncertainty) and irreducible variation in the data (\emph{e.g.}, variation across different sites or scanners). More specifically, elements in $diag(\mathbf{V}^*)$ can be rearranged into the predictive variance tensor $\mathcal{V}^* \in \mathbb{R}^{N^* \times T_1 \times \dots \times T_D}$ reflecting the epistemic uncertainty in predictions . On the other hand, elements in $diag(\mathbf{\Lambda}_i\mathbf{\Sigma}_i\mathbf{\Lambda}_i^\top)$ can be rearranged into a tensor $\mathcal{U} \in \mathbb{R}^{T_1 \times \dots \times T_D}$ reflecting aleatoric uncertainty.

\subsubsection{Efficient Prediction and Optimization} 
\label{subsubsec:optimization}
For efficient prediction and fast optimization of the log-likelihood, we extend the efficient optimization and prediction procedures proposed in~\citet{rakitsch2013all} to cope with our reduced latent space formulations. To this end, we exploit properties of Kronecker product and the eigenvalue decomposition for diagonalizing the covariance matrices in the reduced latent space. Based on our assumption on the orthogonality of components in $\mathbf{B}_i$, we set $\mathbf{B}_i^{-1}=\mathbf{B}_i^\top$ and $\mathbf{B}_i^\top \mathbf{B}_i=\mathbf{I}$ (equivalently for $\mathbf{\Lambda}_i$), in sequel, the predictive mean and variance can be efficiently computed by (see supplementary):
\scriptsize
\begin{subequations} \label{eq:efficient_prediction}
\begin{align}
\mathbf{M}_{(1)}^* = & \mathbf{R^* U_\Omega S_\Omega^{-0.5} U_{\tilde{R}} \tilde{Y} \bigotimes_{i=1}^D U^\top_{\tilde{C}_i} S^{-0.5}_{\Sigma_i} U^{\top}_{\Sigma_i} \Lambda^\top_i B_i C_iB_i^\top}, \\
\begin{split}
\mathbf{V}^*=  & \mathbf{(\bigotimes_{i=1}^D B_iC_iB_i^\top \otimes R^{**}) - (\bigotimes_{i=1}^D B_iC_iB_i^\top \Lambda_i U_{\Sigma_i} S_{\Sigma_i}^{-0.5} U_{\tilde{C}_i}} \\
& \otimes \mathbf{R^* U_\Omega S_{\Omega}^{-0.5} U_{\tilde{R}})} \mathbf{(\bigotimes_{i=1}^D S_{\tilde{C}_i} \otimes S_{\tilde{R}} + I)^{-1}} \\
& \mathbf{(\bigotimes_{i=1}^D U^\top_{\tilde{C}_i} S^{-0.5}_{\Sigma_i} U^{\top}_{\Sigma_i} \Lambda^\top_i B_i C_iB_i^\top \otimes U_{\tilde{R}}^\top S_{\Omega}^{-0.5} U_{\Omega}^\top R^{*\top})},
\end{split}
\end{align}
\end{subequations}
\normalsize
\noindent where in Eq.~\ref{eq:efficient_prediction}a and \ref{eq:efficient_prediction}b we have:
\scriptsize
\begin{align*}
\begin{split}
& vec(\mathbf{\tilde Y}) = diag[\mathbf{(\bigotimes_{i=1}^D S_{\tilde{C}_i} \otimes S_{\tilde{R}} + I)^{-1}}] \odot vec(\mathbf{Y'}), \\
& \mathbf{Y'} =  \mathbf{U_{\tilde{R}}^\top S_{\Omega}^{-0.5} U_{\Omega}^{\top} Y_{(1)} \bigotimes_{i=1}^D \Lambda_i U_{\Sigma_i} S_{\Sigma_i}^{-0.5} U_{\tilde{C}_i}}, \\
& \mathbf{\tilde{C}_i} = \mathbf{S^{-0.5}_{\Sigma_i} U^{\top}_{\Sigma_i} \Lambda^\top_i B_i C_iB_i^\top \Lambda_i U_{\Sigma_i} S^{-0.5}_{\Sigma_i}}, \\ 
& \mathbf{\tilde{R}} = \mathbf{S^{-0.5}_{\Omega} U^{\top}_{\Omega} R U_{\Omega} S^{-0.5}_{\Omega}}. 
\end{split}
\end{align*}
\normalsize
Here $\mathbf{\Sigma_i=U_{\Sigma_i} S_{\Sigma_i} U_{\Sigma_i} ^\top}$ and $\mathbf{\Omega=U_\Omega S_\Omega U_\Omega^\top}$ are eigenvalue decomposition of covariance matrices (similar for $\mathbf{\tilde{C}}_i$ and $\mathbf{\tilde{R}}$). Note that in the new parsimonious formulation for the prediction mean, heavy time and space complexities of computing the inverse kernel matrix is reduced to computing the inverse of a diagonal matrix, \emph{i.e.}, reciprocals of diagonal elements of $\mathbf{\bigotimes_{i=1}^D S_{\tilde{C}_i} \otimes S_{\tilde{R}} + I}$. For the predictive variance, explicit computation of the Kronecker product is still necessary but the required time and storage can be significantly reduced by computing only diagonal members of $\mathbf{V}^*$ in mini-batches. 

To efficiently evaluate the negative log-marginal likelihood of Eq.~\ref{eq:sMT-GPTR}, we have (see supplement for derivation): 
\scriptsize
\begin{align}  \label{eq:LML}
\begin{split} 
L = & -\frac{NT}{2} \ln(2\pi) - \frac{N}{2} \sum_{j=1}^{T} (\ln \bigotimes_{i=1}^D \mathbf{S_{\Sigma_i}})[j,j] 
 - \frac{T}{2} \sum_{j=1}^{N} (\ln \mathbf{S_\Omega}[j,j]) \\ 
& - \frac{1}{2}  \sum_{k=1}^T \sum_{j=1}^N \ln(\bigotimes_{i=1}^D \mathbf{S_{\tilde C_i}}[k,k] \mathbf{S_{\tilde R}}[j,j]+1) \\
& - \frac{1}{2} vec(\mathbf{Y'})^\top (\bigotimes_{i=1}^D \mathbf{S_{\tilde{C}_i} \otimes S_{\tilde{R}} + I})^{-1} vec(\mathbf{Y'})
\end{split}
\end{align}
\normalsize
The proposed sMT-GPTR model has four sets of parameters: 1) $\Theta_{\mathbf{C}_i} \mid_{i=1}^D$, 2) $\Theta_{\mathbf{\Sigma}_i} \mid_{i=1}^D$, 3) $\Theta_{\mathbf{R}}$, and 4) $\Theta_{\mathbf{\Omega}}$; which are optimized by maximizing Eq.\ref{eq:LML} (see supplementary for expressions of relevant gradients). In addition, it has two sets of hyperparameters: 1) $P_i \mid_{i=1}^D$, and 2) $Q_i \mid_{i=1}^D$; that respectively decide the number of components in $\mathbf{B}_i\mid_{i=1}^D$ and $\mathbf{\Lambda}_i\mid_{i=1}^D$. These hyperparameters should be set by means of model selection. 

\subsubsection{Computational Complexities} 
\label{subsubsec:complexity}
The time and space complexities of the proposed method in the optimization phase are $\mathcal{O}(N^3+\sum_{i=1}^{D} P_i^3 + \sum_{i=1}^{D} Q_i^3 + N T^2 + N^2 T)$ and $\mathcal{O}(N^2+\sum_{i=1}^{D} P_i^2+\sum_{i=1}^{D} Q_i^2 + N T)$, respectively. The first three terms belong to the eigenvalue decomposition of $\mathbf{R}$, $\mathbf{\Omega}$, $\mathbf{\Sigma}_i$, and $\mathbf{C}_i$. The last two terms are related to the transformation of $\mathbf{Y_{(1)}}$ to $\mathbf{Y'}$ in Eq.~\ref{eq:LML}. For reasonably small $P_i$ and $Q_i$; and for a very large output space where $T \gg N$, the time and space complexities reduce to $\sim \mathcal{O}(T^2)$ and $\sim \mathcal{O}(T)$ which is one order of magnitude less than the original GP-Kronsum algorithm ($\mathcal{O}(T^3)$ and $\mathcal{O}(T^2)$)~\citep{rakitsch2013all}. Such an improvement yields substantial speed up in the case of neuroimaging data where $T$ is generally in order of $10^5$ or larger. Furthermore, due to the reasonable memory requirement, it makes the impossible mission of multi-task GPR on the whole-brain data possible.

\subsection{Multivariate Normative Modeling} 
\label{subsec:normative_modeling}
As briefly discussed in Sec.~\ref{sec:introduction}, in mass-univariate normative modeling~\citep{marquand2016understanding} single-task GPR (ST-GPR) is employed to independently regress neuroimaging measures from clinical covariates. Thus, it is unable in modeling multivariate signal and noise structures in the neuroimaging data. The proposed sMT-GPTR approach in the TGPP framework provides all the ingredients needed for modeling the multi-way structured variation via \emph{multivariate} normative modeling. Let $\mathcal{Y}^*=\hat{\mathcal{Y}}+\mathcal{M}^* \in \mathbb{R}^{N^* \times T_1 \times \dots \times T_D}$ to represent the predicted neuroimaging data in the TGPP framework using Eq.~\ref{eq:TGPP}. By extending the formulation in~\citet{marquand2016understanding} for computing the normative probability maps to \emph{structured} normative probability maps (S-NPMs) $\mathcal{N} \in \mathbb{R}^{N^* \times T_1 \times \dots \times T_D}$ we have:
\begin{eqnarray} \label{eq:S-NPM}
\mathcal{N} = \frac{\mathcal{Y} - \mathcal{Y}^*}{\sqrt{\mathcal{S}}} \quad ,
\end{eqnarray}
\noindent where $\mathcal{S}$ represents the sum of epistemic and aleatoric uncertainties, \emph{i.e.}, $\mathcal{V}^*$ and $\mathcal{U}$. For example for the $i$th test subject at the $j,k,l$th voxel in the xyz MRI coordinate system, we have $\mathcal{S}[i,j,k,l]=\mathcal{V}^*[i,j,k,l]+\mathcal{U}[j,k,l]$. This new S-NPM formulation enables us to quantify spatio-temporal structured deviations from the multivariate normative model. 
\begin{table*}[h!]
\centering
\caption{Description and number of parameters and hyperparameters in two benchmarked methods.}
\label{tab:parameters}
\resizebox{0.975\textwidth}{!}{
\begin{tabular}{@{}cccc@{}}
\toprule
\textbf{Method} & \textbf{\begin{tabular}[c]{@{}c@{}}No. \\ Parameters\end{tabular}} & \textbf{\begin{tabular}[c]{@{}c@{}}No. \\ Hyperparameters\end{tabular}} & \textbf{Description} \\ \midrule
\textbf{ST-GPR} & 597800 & - & Parameters:$(\left | \Theta_{\mathbf{R}} \right | + \left | \Theta_{\sigma} \right |) \times T = (4+1)\times 119560$ \\ \midrule
\textbf{sMT-GPTR} & 29 & 6 & \begin{tabular}[c]{@{}c@{}}Parameters: $\left | \Theta_{\mathbf{R}} \right | + \left | \Theta_{\mathbf{\Omega}} \right | + \sum_{i=1}^3 \left | \Theta_{\mathbf{C}_i} \right | + \sum_{i=1}^3 \left | \Theta_{\mathbf{\Sigma}_i} \right | = 4+1+12+12$\\ Hyperparameters: $P_1,P_2,P_3,Q_1,Q_2,Q_3$\end{tabular} \\ \bottomrule
\end{tabular}}
\end{table*}

\section{Experiments and Results} \label{sec:results}
\subsection{Experimental Materials and Setup} \label{subsec:materials_setups}
% Dataset and preprocessing
We apply the proposed framework on a clinical neuroimaging dataset from the UCLA Consortium for Neuropsychiatric Phenomics~\citep{poldrack2016phenome}. The preprocessed data~\citep{gorgolewski2017preprocessed} from 119 healthy subjects; and respectively 49, 39, and 48 individuals with schizophrenia (SCHZ), attention deficit hyperactivity disorder (ADHD), and bipolar disorder (BIPL) were used in our experiments.\footnote{Available through the OpenfMRI project at~\url{https://openfmri.org/dataset/ds000030/}.} We used all covariates ($\mathbf{X}$ with $F=30$) from a screening instrument for psychiatric disorders (the ``General Health Questionnaire"\footnote{See ~\url{https://www.statisticssolutions.com/general-health-questionnaire-ghq/}}) to predict a main task effect contrast from the ``task switching" task that is known to be impaired in many clinical conditions~\citep{poldrack2016phenome}. This can be seen as a normative model encoding a general screening tool for psychiatric problems. We used 3D-contrast volumes ($D=3$) with $3mm \times 3mm \times 4mm$ resulotion derived from the standard fMRI preprocessing pipeline presented in~\citet{gorgolewski2017preprocessed}. We cropped the volumes to the minimal bounding-box of $49 \times 61 \times 40$ voxels ($T_1=49,T_2=61,T_3=40,T=119560$).

We compare sMT-GPTR with single-task GPR (ST-GPR), \emph{i.e.}, our multivariate TGPP framework versus the mass-univariate approach, in terms of their normative modeling accuracy and runtime. Note that the comparison with other multi-task GPR approaches is not possible due to their excessive resource requirements when applied to 119560 output variables. For example, in this case GP-Kronsum~\citep{rakitsch2013all} needs at least 80GB memory for storing the task covariance matrix. 

We evaluate the normative modeling accuracy in a novelty detection scenario where we first train a model on a subset of healthy subjects and then calculate NPMs (or S-NPMs) on a test set of healty subjects and patients. As in~\cite{marquand2016understanding}, we use extreme value statistics to provide a statistical model for the deviations. Specifically, we use a block-maximum approach on the top 1\% values in NPMs and fit these to a generalized extreme value distribution (GEVD)~\citep{davison2015statistics}. Then for a given test sample, we interpret the value of the cumulative distribution function of GEVD as the probability of that sample being an abnormal sample~\citep{roberts2000extreme}. Given these probabilities and actual labels, we evaluate the area under the ROC curve (AUC) to measure the performance of the model in distinguishing between healthy individuals from patients. To this end, we randomly divide the data into three subsets: 1) 39 healthy subjects to train models; 2) 39 healthy subjects to estimate the parameters of the GEVD; and 3) 41 healthy subjects and patients data in the test set. All steps (random sampling, modeling, and evaluation) are repeated 10 times in order to estimate the fluctuations of models trained on different training sets.

% GP settings
In all above experiments, we use ordinary least squares to estimate the fixed-effect in Eq.~\ref{eq:fixed_effect}. In the sMT-GPTR case, the Tucker model~\citep{tucker1966some} from Tensorly package~\citep{kossaifi2016tensorly} is used for tensor factorization in which we set correspondingly $P_1=P_2=P_3=3,5,10,15$ and $Q_1=Q_2=Q_3=1,3,5,10$.\footnote{It is worthwhile to emphasize that the proposed method does not make any assumption on the type of tensor factorization method, thus any other tensor decomposition approaches (such as PARAFAC) can be applied as well.} In all models, we use a composite covariance function of a linear, a squared exponential, and a diagonal isotropic covariance functions for $\mathbf{R}, \mathbf{C}_i \mid_{i=1}^3$, and $\mathbf{\Sigma}_i \mid_{i=1}^3$; and a diagonal isotropic covariance function for $\mathbf{\Omega}$. The truncated Newton algorithm is used for optimizing the parameters. Table~\ref{tab:parameters} summarizes the number of parameters and hyperparameters of two benchmarked methods. All experiments are performed using an Intel\textsuperscript \textregistered Xeon\textsuperscript \textregistered E5-2640 v3 @2.60GHz CPU and 16GB of RAM.\footnote{Implementations are made available online at~\url{www.anonymous.link}.}

% Supervised SVM setting
We further compare the unsupervised normative modeling approach with an off-the-shelf support vector machine (SVM) classifier (as is a standard practice in fMRI) in predicting the diagnostic labels of three different disorders (schizophrenia, ADHD and bipolar disorder). To this end, in a stratified 5-fold cross-validation setting, we evaluated three binary SVM classifiers (\emph{i.e.}, healthy \emph{vs.} SCHZ, healthy \emph{vs.} ADHD, and healthy \emph{vs.} BIPL) in predicting the diagnosis labels from the fMRI data. Here, the main task effect contrasts from the ``task switching'' task are used as input to the SVM classifier. In each cross-validation fold, the grid-search approach on the training set is used to find best kernel among linear and radial basis function (RBF); and the best value for the slack parameter and kernel width (in RBF kernel) among $\{10^{-3},10^{-2},10^{-1},1,10^{1},10^{2},10^{3}\}$.\footnote{The scikit-learn toolbox~\citep{pedregosa2011scikit} is used for training and testing the SVM classifier.}
\begin{figure*}[t]
	\centering
	\includegraphics[width=0.995\textwidth]{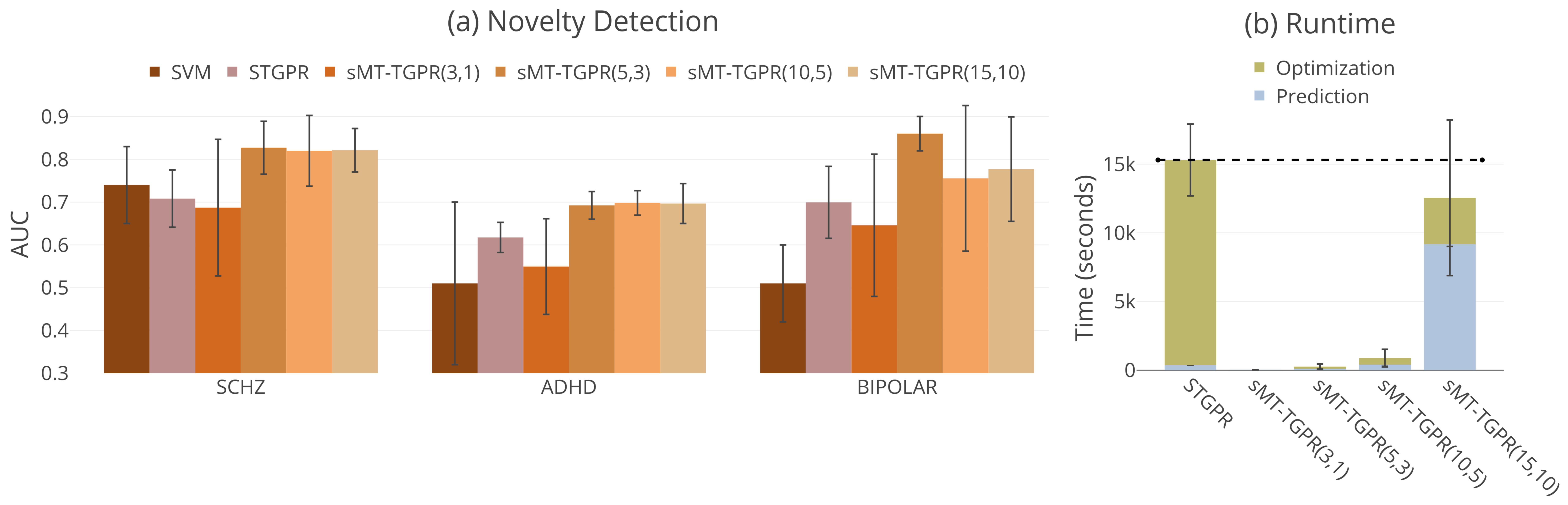}
	\caption{Comparison between ST-GPR and sMT-GPTR in terms of: a) AUC in the abnormal sample detection using normative modeling, and b) optimization and prediction runtime. The numbers in the parentheses show the number of components used in tensor factorization of $\hat{\mathcal{Z}}$ and $\hat{\mathcal{E}}$.}
	\label{fig:Comparison_bar_plots}
\end{figure*}

\subsection{sMT-GPTR: Faster, More Accurate, and Feasible in Whole-Brain Inference} \label{subsec:results_comparison}
Figure~\ref{fig:Comparison_bar_plots} compares the AUC and runtime of ST-GPR with those of sMT-GPTR for different numbers of components in tensor factorization. As illustrated in Figure~\ref{fig:Comparison_bar_plots}(a), accounting for spatial structures of the signal and noise in the multi-task learning setting provides normative models with better detection accuracy relative to single-task learning. Using sufficient components in the tensor factorization, the sMT-GPTR approach provides substantially higher accuracy in detecting abnormal samples across all diagnosis labels. Considering the fact that ST-GPR and sMT-GPTR models showed similar regression performance (see supplement), the AUC boost in sMT-GPTR models probably reflect better estimations of epistemic and aleatoric uncertainties. Our results show that using sMT-GPTR with 5 and 3 components to respectively explain the variances of the random-effect and noise is enough to reach the highest detection accuracy.

The gain in the detection accuracy is even more pronounced in comparison with the supervised SVM classifier. SVM achieves inferior AUC ($0.74 \pm 0.09$) compared to our unsupervised approach in classifying SCHZ patients and its performance remains at the chance-level in ADHD and BIPL cases. The fact that our approach outperforms a fully supervised approach despite never having seen a patient indicates that the target pattern is not consistent across individuals within the patient group~\citep{wolfers2015estimating}. Instead, the normative model focuses only on estimating the healthy distribution and can detect differences from this distribution regardless of whether they are consistent with one another. Moreover, in the supervised scenario, even though the model has access to labels, it cannot benefit from the information in the covariates. While in the normative modeling framework both sources of information (in covariates and fMRI data) are exploited. 

In addition to making multi-task learning possible in a very high-dimensional setting, for a reasonable number of components, sMT-GPTR is significantly faster than ST-GPR in terms of total runtime (Fig.~\ref{fig:Comparison_bar_plots}(b)). For example, sMT-GPTR(10,5) is 17 times faster than ST-GPR reducing its runtime from $\sim 4$ hours to $\sim 15$ minutes. Even though the model selection process to decide the number of components is a time-consuming step in practice, due to the low running time of the proposed approach, it remains economical compared to other multi-task alternatives. It the end, it is worthwhile to emphasize that these improvements are achieved by reducing the degree-of-freedom of the normative model from 597800 for ST-GPR to $29+6=35$ for sMT-GPTR (see Table~\ref{tab:parameters} for the number of parameters and hyperparameters of different models). 
\begin{figure*}[t]
	\centering
	\includegraphics[width=0.995\textwidth]{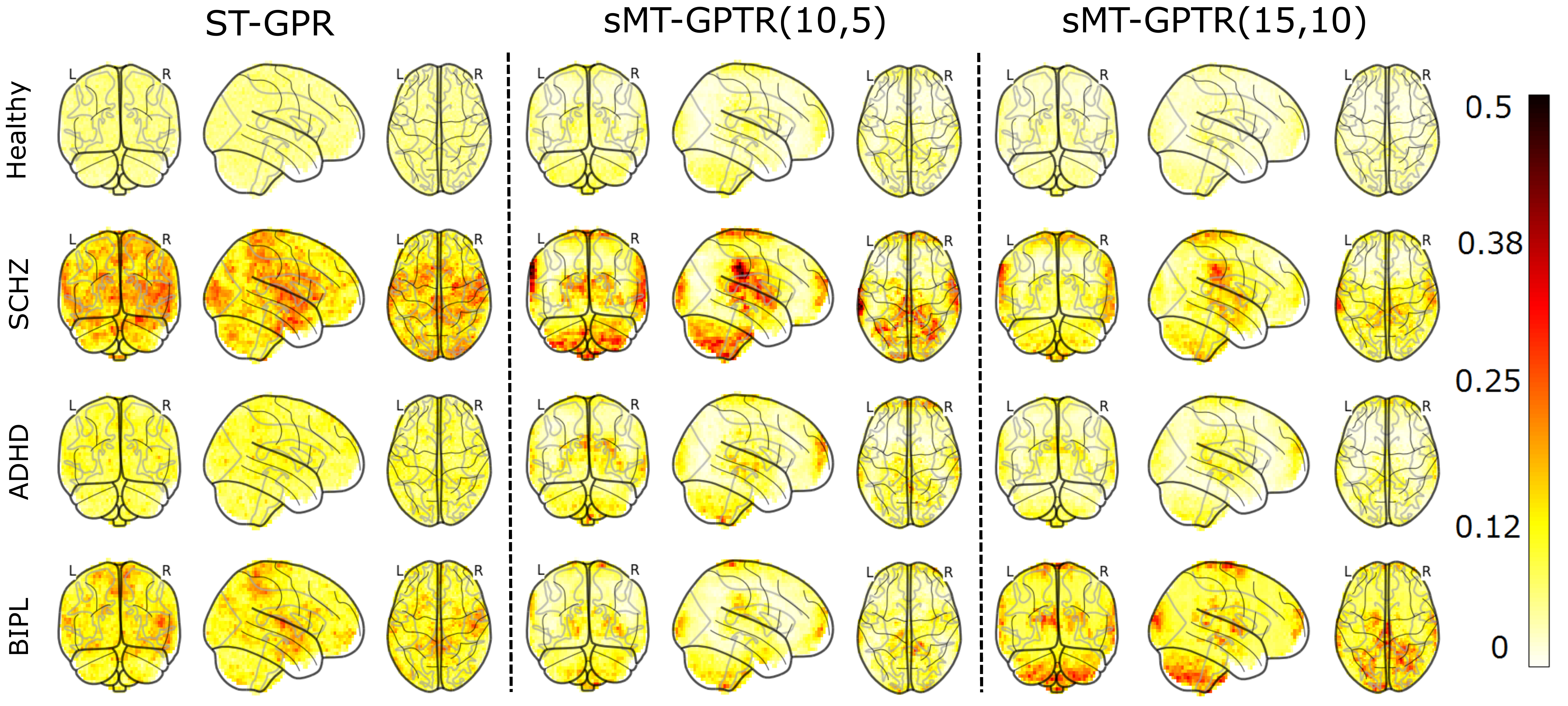}
	\caption{The probability of each voxel to deviate from the normative model in healthy and patient populations; derived by ST-GPR, sMT-GPTR(10,5), and sMT-GPTR(15,10).}
	\label{fig:Brain_plots}
\end{figure*}
	
\subsection{Understanding the Underlying Neural Patterns of Abnormality} 
\label{subsec:results_brainplots}
We have shown that accounting for spatial structure provides more accurate normative models than the baseline single-task model. However, it is also important to understand the neural basis of the underlying abnormalities. To achieve this for ST-GPR and sMT-GPTR, we use a spatial mixture model~\citep{woolrich2005mixture} to translate the corresponding NPM and S-NPM of each subject to a probability map, where the value of each voxel represents the probability that voxel deviates from the normative model~\citep{wolfers2016quantifying}. Figure~\ref{fig:Brain_plots} shows the resulting probability maps for ST-GPR, sMT-GPTR(10,5), and sMT-GPTR(15,10) averaged across runs and the healthy/patients population in the test set.\footnote{Plots are created using the Nilearn toolbox~\citep{abraham2014machine}. See supplementary for probability maps of sMT-GPTR(3,1) and sMT-GPTR(5,3).} These maps illustrate that: i) in general the probability of deviating from the normative model is higher in patients than healthy subjects. These deviations are more salient in SCHZ and BIPL patients compared to ADHD patients. This obseravation is compatible with higher novelty detection performance in SCHZ and BIPL patients (see Figure~\ref{fig:Comparison_bar_plots}(a)); ii) the areas with high deviation probability are more spatially focal in sMT-GPTR models than the ST-GPR model. This suggests that the sMT-GPTR approach is better able to focus on the core abnormalities underlying the disorder and that accounting for spatial structure in both random-effect and noise provides a better estimation of the structured epistemic and aleatoric uncertainties in sMT-GPTR compared to ST-GPR models.
\section{Related Work} 
\label{sec:related_works}
\citet{hyun2014SGPP,hyun2016STGP} introduced spatial and spatio-temporal Gaussian predictive process to model neuroimaging data. They used functional principal component analysis to approximate the spatial/temporal covariance matrix of the random-effect combined with a multivariate autoregressive model for the noise. Their approach focuses on point estimation of outputs and does not provide a practical solution to estimate predictive uncertainty, thus cannot be employed for normative modeling. Our TGPP framework resolves this issue, and further, due to its flexible and general tensor assumption on the data structure, can be extended to other possible dimensions of neuroimaging data in addition to space and time.  

\citet{shvartsman2017matrix} reformulated common fMRI analysis methods, such as representational similarity analysis, using matrix-variate normal formalism resulting in a unified framework for fMRI data analysis. They theoretically and experimentally showed the potentials of matrix-normal assumption on fMRI data in simultaneously modeling spatial and temporal noise covariances. Although our aim is different, our tensor-variate normal assumption on the distribution of the random-effect and noise can be seen as an extension of their approach, extending theoretical concepts in the multi-way modeling of neuroimaging data from 2-dimensional matrix-structured to $D$-dimensional tensor-structured data.   

Exploiting the properties of Kronecker algebra to scale up the computational complexities of GPR in analyzing multi-way data is well studied in machine learning literature~\citep{saatcci2012scalable,wilson2014fast,wilson2015kernel,gilboa2015scaling,izmailov2018scalable}. However, all studies in this direction are mainly focus on multi-way \emph{input} space, (\emph{i.e.}, single-task GPR), whereas we extend this ideas to multi-way \emph{output} space, (\emph{i.e.}, multi-task GPR). This extension is one of our core contributions that makes the multivariate normative modeling possible.

The idea of using a sum of Kronecker products as the covariance term in order to concurrently learn structured signal and noise covariance functions in a multi-task Gaussian process setting is introduced first time in~\citet{rakitsch2013all}, known as GP-Kronsum. We have extended their method from two important perspectives: i) MT-GPTR generalizes the core idea of learning structured signal and noise covariance matrices to D-dimensional multi-way tensor structured data. This generalization not only provides the possibility of learning more complex multi-way structures but also reduces the computational complexities of GP-Kronsum by utilizing a more fine-grained Kronecker structure across different tensor dimensions; ii) we analytically show how using tensor factorization technique for low-rank approximation of covariance matrices can respectively decrease the time and space complexity of GP-Kronsum from $\mathcal{O}(T^3)$ and $\mathcal{O}(T^2)$ to $\mathcal{O}(T^2)$ and $\mathcal{O}(T)$, \emph{i.e.}, one order of magnitude improvement. These massive improvements are crucial especially for applications on high-dimensional neuroimaging data.
\section{Summary, Limitation, and Future Work} \label{sec:conclusions}
In this study, assuming a tensor-variate normal distribution on multi-way neuroimaging data and in a novel tensor Gaussian predictive process framework, we introduced a scalable multi-task Gaussian process tensor regression approach to model multi-way structured random-effect and noise on very high-dimensional neuroimaging data. The proposed approach provides a breakthrough toward practical modeling different sources of variations across different dimensions of large neuroimaging cohorts. On a clinical fMRI dataset, we exemplified one possible application of the proposed method for multivariate normative modeling of spatially distributed effects at the whole-brain level. We demonstrated that our framework provides more accurate results with reasonable computational costs, and it focuses better on the core underlying brain abnormalities relative to its mass-univariate alternative. 

Due to its tensor-based design, the presented TGPP framework needs full-grid data across space, and/or other possible dimensions of neuroimaging data. This can be considered as a possible limitation when dealing with data with missing values across some data dimensions. One possible future direction is to solve this problem by imputing the grid using imaginary observations~\citep{wilson2014fast,wilson2015kernel}. For future work, we aim to better understand the neuroscientific basis for the performance improvements we report (\emph{e.g.}, across multiple model orders and using different representations of the normative probability maps) and will apply the proposed method to very large cohorts in order to provide a more comprehensive model of biological variation in human brain.
\bibliography{references}
\onecolumn
\newpage
\section*{Supplementary Materials}
\label{sec:supplementary}
Throughout the supplementary materials we use the same notation introduced in the main text.

\subsection*{Useful Equations}
\label{subsec:useful}
For $\mathbf{A} \in \mathbb{R}^{M \times N}$, $\mathbf{B} \in \mathbb{R}^{P \times Q}$, and $\mathbf{C}$, $\mathbf{D}$ (with appropriate size) we have:
\begin{enumerate}
\item $\mathbf{A = U_A S_A U_A^\top}$ is the eigenvalue decomposition of $\mathbf{A}$,
\item $\mathbf{(ACB)^{-1}=B^{-1}C^{-1}A^{-1}}$,
\item $\mathbf{(A \otimes B)(C \otimes D) = AC \otimes BD}$,
\item $\mathbf{(A \otimes B)^{-1} = A^{-1} \otimes B^{-1}}$,
\item the eigenvalue decomposition of $\mathbf{A \otimes B + I}$ is: $\mathbf{(U_A \otimes U_B)(S_A \otimes S_B + I)(U_A^\top \otimes U_B^\top)}$,
\item $(\mathbf{A \otimes B}) vec(\mathbf{C}) = vec(\mathbf{BCA}^{\top})$,
\item $\ln \left | \mathbf{AC} \right | = \ln (\left | \mathbf{A} \right | \left | \mathbf{C} \right |) = \ln \left | \mathbf{A} \right | + \ln \left | \mathbf{C} \right |$,
\item for $\mathbf{C} \in \mathbb{R}^{N \times N}$, $\frac{\mathrm{d}}{\mathrm{d} x} \ln \left | \mathbf{C} \right | = Tr[\mathbf{C}^{-1} \frac{\mathrm{d} \mathbf{C}}{\mathrm{d} x}]$,
\item $Tr[\mathbf{ACBD}]=Tr[\mathbf{CBDA}]=Tr[\mathbf{BDAC}]=Tr[\mathbf{DACB}]$,
\item $Tr[\mathbf{A^{\top}C}]=vec(C)^{\top}vec(A)$,
\item $\ln \left | \mathbf{C} \otimes \mathbf{D} \right | = M \ln \left | \mathbf{C} \right | + N \ln \left | \mathbf{D} \right |, \quad for \quad \mathbf{C} \in \mathbb{R}^{N \times N}, \mathbf{D} \in \mathbb{R}^{M \times M}$,
\item $\ln \left | diag(\mathbf{C}) \right | = \prod_{i=1}^N \mathbf{C}[i,i], \quad for \quad \mathbf{C} \in \mathbb{R}^{N \times N}$.
\end{enumerate}

\subsection*{Tensor Normal Distribution for $\mathcal{Z}'$}
\label{subsec:mean_prediction}
Eq.~\ref{eq:TND_Zprime} is derived as follows:
\small
\begin{eqnarray*} \label{eq:z_prime}
\begin{split}
& p(\mathcal{Z'}  \mid \mathbf{C}_i \mid_{i=1}^D, \mathbf{R}) = \mathcal{TN}(\mathbf{0}, \bigotimes_{i=1}^D \mathbf{C}_i \otimes \mathbf{R}) = \frac{\exp(-\frac{1}{2}vec(\mathcal{Z'})^\top [\bigotimes_{i=1}^D \mathbf{C}_i \otimes \mathbf{R}]^{-1}vec(\mathcal{Z'}))}{\sqrt{(2\pi)^{N \prod_{i=1}^D P_i} \left | \bigotimes_{i=1}^D \mathbf{C}_i \right |^{N} \left | \mathbf{R} \right |^{\prod_{i=1}^D P_i}}} = \\
& \frac{\exp(-\frac{1}{2}vec(\mathcal{Z'})^\top vec(\mathbf{R}^{-1} \mathbf{Z'_{(1)}}\bigotimes_{i=1}^D \mathbf{C}_i^{-1}))}{\sqrt{(2\pi)^{N \prod_{i=1}^D P_i} \left | \bigotimes_{i=1}^D \mathbf{C}_i \right |^{N} \left | \mathbf{R} \right |^{\prod_{i=1}^D P_i}}} =
\frac{\exp(-\frac{1}{2} Tr[\bigotimes_{i=1}^D \mathbf{C}_i^{-1} \mathbf{Z'^{\top}_{(1)} \mathbf{R}^{-1} \mathbf{Z'}_{(1)}}])}{\sqrt{(2\pi)^{N \prod_{i=1}^D P_i} \left | \bigotimes_{i=1}^D \mathbf{C}_i \right |^{N} \left | \mathbf{R} \right |^{\prod_{i=1}^D P_i}}} \\
& \xrightarrow[]{\mathbf{Z'_{(1)}}=\mathbf{\hat Z_{(1)}}(\bigotimes_{i=1}^D \mathbf{B}_i^{\top})^{-1}} 
\frac{\exp(-\frac{1}{2} Tr[\bigotimes_{i=1}^D \mathbf{C}_i^{-1} \bigotimes_{i=1}^D \mathbf{B}_i^{\dagger} \mathbf{\hat Z_{(1)}}^{\top} \mathbf{R}^{-1} \mathbf{\hat Z_{(1)}}(\bigotimes_{i=1}^D \mathbf{B}_i^{\top})^{-1}])}{\sqrt{(2\pi)^{N \prod_{i=1}^D P_i} \left | \bigotimes_{i=1}^D \mathbf{C}_i \right |^{N} \left | \mathbf{R} \right |^{\prod_{i=1}^D P_i}}} = \\
& \frac{\exp(-\frac{1}{2} Tr[\bigotimes_{i=1}^D \mathbf{B}_i\mathbf{C}_i^{-1}\mathbf{B}_i^{\top} \mathbf{\hat Z_{(1)}}^{\top} \mathbf{R}^{-1} \mathbf{\hat Z_{(1)}}])}{\sqrt{(2\pi)^{N \prod_{i=1}^D P_i} \left | \bigotimes_{i=1}^D \mathbf{C}_i \right |^{N} \left | \mathbf{R} \right |^{\prod_{i=1}^D P_i}}} \quad .
\end{split}
\end{eqnarray*}
\normalsize

\subsection*{Efficient Mean Prediction}
\label{subsec:mean_prediction}
Eq.~\ref{eq:efficient_prediction}(a) is derived from Eq.~\ref{eq:predictive_distribution}(a) as follows:
\small
\begin{eqnarray*} \label{eq:eff_mean_pred}
\begin{split}
vec(\mathcal{M}^*) = & (\bigotimes_{i=1}^D \mathbf{B}_i\mathbf{C}_i\mathbf{B}_i^\top \otimes \mathbf{R}^*)(\bigotimes_{i=1}^D \mathbf{B}_i\mathbf{C}_i\mathbf{B}_i^\top \otimes \mathbf{R} + \bigotimes_{i=1}^D \mathbf{\Lambda}_i\mathbf{\Sigma}_i\mathbf{\Lambda}_i^\top \otimes \mathbf{\Omega})^{-1} vec(\mathcal{Y}) 
\\
= & (\bigotimes_{i=1}^D \mathbf{B}_i\mathbf{C}_i\mathbf{B}_i^\top \otimes \mathbf{R}^*)(\bigotimes_{i=1}^D \mathbf{B}_i\mathbf{C}_i\mathbf{B}_i^\top \otimes \mathbf{R} + \bigotimes_{i=1}^D \mathbf{\Lambda}_i\mathbf{U_{\Sigma_i}S_{\Sigma_i}U_{\Sigma_i}^{\top}}\mathbf{\Lambda}_i^\top \otimes \mathbf{U_{\Omega}S_{\Omega}U_{\Omega}^{\top}})^{-1} vec(\mathcal{Y}) \\
= & (\bigotimes_{i=1}^D \mathbf{B}_i\mathbf{C}_i\mathbf{B}_i^\top \otimes \mathbf{R}^*)(\bigotimes_{i=1}^D \mathbf{\Lambda}_i\mathbf{U_{\Sigma_i}S_{\Sigma_i}^{-0.5}} \otimes \mathbf{U_{\Omega}S_{\Omega}^{-0.5}})(\bigotimes_{i=1}^D \mathbf{\tilde{C}_i} \otimes \mathbf{\tilde{R}} + \mathbf{I})^{-1} \\ 
& (\bigotimes_{i=1}^D \mathbf{S_{\Sigma_i}^{-0.5}U_{\Sigma_i}^{\top}\Lambda_i^{\top}} \otimes \mathbf{S_{\Omega}^{-0.5}U_{\Omega}^{\top}})vec(\mathcal{Y}) \\
= & (\bigotimes_{i=1}^D \mathbf{B}_i\mathbf{C}_i\mathbf{B}_i^\top \mathbf{\Lambda}_i\mathbf{U_{\Sigma_i}S_{\Sigma_i}^{-0.5}} \otimes \mathbf{R}^*\mathbf{U_{\Omega}S_{\Omega}^{-0.5}})(\bigotimes_{i=1}^D \mathbf{U_{\tilde C_i}} \otimes \mathbf{U_{\tilde{R}}})(\bigotimes_{i=1}^D \mathbf{S_{\tilde{C}_i}} \otimes \mathbf{S_{\tilde{R}}}+\mathbf{I})^{-1}\\ 
& (\bigotimes_{i=1}^D \mathbf{U_{\tilde{C}_i}^{\top}} \otimes \mathbf{U_{\tilde{R}}^{\top}})  
vec(\mathbf{S_{\Omega}^{-0.5}U_{\Omega}^{\top} Y_{(1)}} \bigotimes_{i=1}^D \mathbf{\Lambda}_i\mathbf{U_{\Sigma_i}S_{\Sigma_i}^{-0.5}}) \\
= & (\bigotimes_{i=1}^D \mathbf{B}_i\mathbf{C}_i\mathbf{B}_i^\top \mathbf{\Lambda}_i\mathbf{U_{\Sigma_i}S_{\Sigma_i}^{-0.5}} \otimes \mathbf{R}^*\mathbf{U_{\Omega}S_{\Omega}^{-0.5}})(\bigotimes_{i=1}^D \mathbf{U_{\tilde{C}_i}} \otimes \mathbf{U_{\tilde{R}}}) vec(\mathbf{\tilde Y}) \\
= & (\bigotimes_{i=1}^D \mathbf{B}_i\mathbf{C}_i\mathbf{B}_i^\top \mathbf{\Lambda}_i\mathbf{U_{\Sigma_i}S_{\Sigma_i}^{-0.5}} \otimes \mathbf{R}^*\mathbf{U_{\Omega}S_{\Omega}^{-0.5}}) vec(\mathbf{U_{\tilde{R}}} \mathbf{\tilde Y} \bigotimes_{i=1}^D \mathbf{U_{\tilde{C}_i}^{\top}}) \\
= & \mathbf{R^* U_\Omega S_\Omega^{-0.5} U_{\tilde{R}} \tilde{Y} \bigotimes_{i=1}^D U^\top_{\tilde{C}_i} S^{-0.5}_{\Sigma_i} U^{\top}_{\Sigma_i} \Lambda^\top_i B_i C_iB_i^\top}
\quad .
\end{split}
\end{eqnarray*}
\normalsize

\subsection*{Efficient Variance Prediction}
\label{subsec:variance_prediction}
Eq.~\ref{eq:efficient_prediction}(b) is derived from Eq.~\ref{eq:predictive_distribution}(b) as follows:
\small
\begin{eqnarray*} \label{eq:eff_var_pred}
\begin{split}
\mathbf{V}^* = &  (\bigotimes_{i=1}^D \mathbf{B}_i\mathbf{C}_i\mathbf{B}_i^\top \otimes \mathbf{R}^{**})-(\bigotimes_{i=1}^D \mathbf{B}_i\mathbf{C}_i\mathbf{B}_i^\top \otimes \mathbf{R}^*) (\bigotimes_{i=1}^D \mathbf{B}_i\mathbf{C}_i\mathbf{B}_i^\top \otimes \mathbf{R} + \bigotimes_{i=1}^D \mathbf{\Lambda}_i\mathbf{\Sigma}_i\mathbf{\Lambda}_i^\top \otimes \mathbf{\Omega})^{-1}\\ & (\bigotimes_{i=1}^D \mathbf{B}_i\mathbf{C}_i\mathbf{B}_i^\top \otimes \mathbf{R}^{* \top}) \\
= &  (\bigotimes_{i=1}^D \mathbf{B}_i\mathbf{C}_i\mathbf{B}_i^\top \otimes \mathbf{R}^{**})-(\bigotimes_{i=1}^D \mathbf{B}_i\mathbf{C}_i\mathbf{B}_i^\top \otimes \mathbf{R}^*)(\bigotimes_{i=1}^D \mathbf{\Lambda}_i\mathbf{U_{\Sigma_i}S_{\Sigma_i}^{-0.5}} \otimes \mathbf{U_{\Omega}S_{\Omega}^{-0.5}}) \\
& (\bigotimes_{i=1}^D \mathbf{\tilde{C}_i} \otimes \mathbf{\tilde{R}} + \mathbf{I})^{-1}(\bigotimes_{i=1}^D \mathbf{S_{\Sigma_i}^{-0.5}U_{\Sigma_i}^{\top}\Lambda_i^{\top}} \otimes \mathbf{S_{\Omega}^{-0.5}U_{\Omega}^{\top}})(\bigotimes_{i=1}^D \mathbf{B}_i\mathbf{C}_i\mathbf{B}_i^\top \otimes \mathbf{R}^{* \top}) \\
= &  (\bigotimes_{i=1}^D \mathbf{B}_i\mathbf{C}_i\mathbf{B}_i^\top \otimes \mathbf{R}^{**})-(\bigotimes_{i=1}^D \mathbf{B}_i\mathbf{C}_i\mathbf{B}_i^\top \mathbf{\Lambda}_i\mathbf{U_{\Sigma_i}S_{\Sigma_i}^{-0.5}} \otimes \mathbf{R}^* \mathbf{U_{\Omega}S_{\Omega}^{-0.5}}) \\
& (\bigotimes_{i=1}^D \mathbf{U_{\tilde C_i}} \otimes \mathbf{U_{\tilde{R}}})(\bigotimes_{i=1}^D \mathbf{S_{\tilde{C}_i}} \otimes \mathbf{S_{\tilde{R}}}+\mathbf{I})^{-1}(\bigotimes_{i=1}^D \mathbf{U_{\tilde{C}_i}^{\top}} \otimes \mathbf{U_{\tilde{R}}^{\top}})(\bigotimes_{i=1}^D \mathbf{S_{\Sigma_i}^{-0.5}U_{\Sigma_i}^{\top}\Lambda_i^{\top}} \mathbf{B}_i\mathbf{C}_i\mathbf{B}_i^\top \otimes \mathbf{S_{\Omega}^{-0.5}U_{\Omega}^{\top}}\mathbf{R}^{* \top}) \\
= & \mathbf{(\bigotimes_{i=1}^D B_iC_iB_i^\top \otimes R^{**})- (\bigotimes_{i=1}^D B_iC_iB_i^\top \Lambda_i U_{\Sigma_i} S_{\Sigma_i}^{-0.5} U_{\tilde{C}_i} \otimes R^* U_\Omega S_{\Omega}^{-0.5} U_{\tilde{R}})} \\
& \mathbf{(\bigotimes_{i=1}^D S_{\tilde{C}_i} \otimes S_{\tilde{R}} + I)^{-1} (\bigotimes_{i=1}^D U^\top_{\tilde{C}_i} S^{-0.5}_{\Sigma_i} U^{\top}_{\Sigma_i} \Lambda^\top_i B_i C_iB_i^\top \otimes U_{\tilde{R}}^\top S_{\Omega}^{-0.5} U_{\Omega}^\top R^{*\top})}
\quad .
\end{split}
\end{eqnarray*}
\normalsize

\subsection*{Efficient Log Marginal Likelihood Evaluation}
\label{subsec:LML_evaluation}
Eq.~\ref{eq:LML} is derived as follows:
\small
\begin{eqnarray*} \label{eq:eff_LML}
\begin{split}
& L = -\frac{NT}{2} \ln(2\pi)-\frac{1}{2}\ln\left | \mathbf{K} \right | - \frac{1}{2} vec(\mathbf{Y})^\top \mathbf{K}^{-1} vec(\mathbf{Y}) \\
& = -\frac{NT}{2} \ln(2\pi)-\frac{1}{2}\ln\left |\bigotimes_{i=1}^D \mathbf{B}_i\mathbf{C}_i\mathbf{B}_i^\top \otimes \mathbf{R} + \bigotimes_{i=1}^D \mathbf{\Lambda}_i\mathbf{\Sigma}_i\mathbf{\Lambda}_i^\top \otimes \mathbf{\Omega} \right | - \\
& \frac{1}{2} vec(\mathbf{Y})^\top (\bigotimes_{i=1}^D \mathbf{B}_i\mathbf{C}_i\mathbf{B}_i^\top \otimes \mathbf{R} + \bigotimes_{i=1}^D \mathbf{\Lambda}_i\mathbf{\Sigma}_i\mathbf{\Lambda}_i^\top \otimes \mathbf{\Omega})^{-1} vec(\mathbf{Y}) \\ 
& = -\frac{NT}{2} \ln(2\pi)-\frac{1}{2}\ln\left |(\bigotimes_{i=1}^D \mathbf{\Lambda}_i\mathbf{U_{\Sigma_i}S_{\Sigma_i}^{-0.5}} \otimes \mathbf{U_{\Omega}S_{\Omega}^{-0.5}})(\bigotimes_{i=1}^D \mathbf{\tilde{C}_i} \otimes \mathbf{\tilde{R}} + \mathbf{I})^{-1}(\bigotimes_{i=1}^D \mathbf{S_{\Sigma_i}^{-0.5}U_{\Sigma_i}^{\top}\Lambda_i^{\top}} \otimes \mathbf{S_{\Omega}^{-0.5}U_{\Omega}^{\top}}) \right | - \\
& \frac{1}{2} vec(\mathbf{Y})^\top [(\bigotimes_{i=1}^D \mathbf{\Lambda}_i\mathbf{U_{\Sigma_i}S_{\Sigma_i}^{-0.5}} \otimes \mathbf{U_{\Omega}S_{\Omega}^{-0.5}})(\bigotimes_{i=1}^D \mathbf{\tilde{C}_i} \otimes \mathbf{\tilde{R}} + \mathbf{I})^{-1}(\bigotimes_{i=1}^D \mathbf{S_{\Sigma_i}^{-0.5}U_{\Sigma_i}^{\top}\Lambda_i^{\top}} \otimes \mathbf{S_{\Omega}^{-0.5}U_{\Omega}^{\top}})] vec(\mathbf{Y}) \\ 
& = -\frac{NT}{2} \ln(2\pi) - \frac{1}{2}\ln\left |\bigotimes_{i=1}^D \mathbf{\Lambda}_i\mathbf{U_{\Sigma_i}S_{\Sigma_i}U_{\Sigma_i}^{\top}}\mathbf{\Lambda}_i^\top \otimes \mathbf{U_{\Omega}S_{\Omega}U_{\Omega}^{\top}} \right | - \frac{1}{2}\ln\left |\bigotimes_{i=1}^D \mathbf{S_{\tilde{C}_i}} \otimes \mathbf{S_{\tilde{R}}}+\mathbf{I} \right | \\
& - \frac{1}{2}  vec(\mathbf{S_{\Omega}^{-0.5}U_{\Omega}^{\top} Y_{(1)}} \bigotimes_{i=1}^D \mathbf{\Lambda}_i\mathbf{U_{\Sigma_i}S_{\Sigma_i}^{-0.5}})^\top (\bigotimes_{i=1}^D \mathbf{U_{\tilde C_i}} \otimes \mathbf{U_{\tilde{R}}})(\bigotimes_{i=1}^D \mathbf{S_{\tilde{C}_i}} \otimes \mathbf{S_{\tilde{R}}}+\mathbf{I})^{-1} \\ 
& (\bigotimes_{i=1}^D \mathbf{U_{\tilde{C}_i}^{\top}} \otimes \mathbf{U_{\tilde{R}}^{\top}}) vec(\mathbf{S_{\Omega}^{-0.5}U_{\Omega}^{\top} Y_{(1)}} \bigotimes_{i=1}^D \mathbf{\Lambda}_i\mathbf{U_{\Sigma_i}S_{\Sigma_i}^{-0.5}}) \\
& = -\frac{NT}{2} \ln(2\pi) - \frac{1}{2}\ln\left |\bigotimes_{i=1}^D \mathbf{S_{\Sigma_i}} \otimes \mathbf{S_{\Omega}} \right | - \frac{1}{2}\ln\left |\bigotimes_{i=1}^D \mathbf{S_{\tilde{C}_i}} \otimes \mathbf{S_{\tilde{R}}}+\mathbf{I} \right | - \frac{1}{2}  vec(\mathbf{Y'})^\top (\bigotimes_{i=1}^D \mathbf{S_{\tilde{C}_i}} \otimes \mathbf{S_{\tilde{R}}}+\mathbf{I})^{-1} vec(\mathbf{Y'}) \\
& = -\frac{NT}{2} \ln(2\pi) - \frac{N}{2} \sum_{j=1}^{T} (\ln \bigotimes_{i=1}^D \mathbf{S_{\Sigma_i}})[j,j] 
 - \frac{T}{2} \sum_{j=1}^{N} (\ln \mathbf{S_\Omega}[j,j])  - \frac{1}{2} \\ 
& \sum_{k=1}^T \sum_{j=1}^N \ln(\bigotimes_{i=1}^D \mathbf{S_{\tilde C_i}}[k,k] \mathbf{S_{\tilde R}}[j,j]+1) - \frac{1}{2} vec(\mathbf{Y'})^\top (\bigotimes_{i=1}^D \mathbf{S_{\tilde{C}_i} \otimes S_{\tilde{R}} + I})^{-1} vec(\mathbf{Y'}) \quad .
\end{split}
\end{eqnarray*}
\normalsize

\subsection*{Derivatives of $L$ with Respect to Parameters}
\label{subsec:gradients}
In the optimization process, the derivatives of $L$ with respect to $\theta_{\mathbf{C}_i} \in \Theta_{\mathbf{C}_i}$, $\theta_{\mathbf{\Sigma}_i} \in \Theta_{\mathbf{\Sigma}_i}$, $\theta_{\mathbf{R}} \in \Theta_{\mathbf{R}}$, and $\theta_{\mathbf{\Omega}} \in \Theta_{\mathbf{\Omega}}$ can be efficiently computed as follows: 

\subsubsection*{Gradients of $L$ with Respect to $\theta_{\mathbf{C}_i}$}
\label{subsubsec:gradient_c}
\scriptsize
\begin{eqnarray*} \label{eq:derivatives_c}
\begin{split}
\frac{\partial L}{\partial \theta_{\mathbf{C}_i}} & = -\frac{1}{2}diag((\bigotimes_{k=1}^D \mathbf{S_{\tilde{C}_k} \otimes S_{\tilde{R}} + I})^{-1})^\top \\
& diag(\mathbf{S_{\tilde{C}_1} \otimes S_{\tilde{C}_2} \otimes \dots \otimes U_{\tilde{C}_i}^{\top} S^{-0.5}_{\Sigma_i} U^{\top}_{\Sigma_i} \Lambda^\top_i B_i \frac{\partial C_i}{\partial \theta_{\tilde{C}_i}} B_i^\top \Lambda_i U_{\Sigma_i} S^{-0.5}_{\Sigma_i} U_{\tilde{C}_i} \otimes \dots \otimes S_{\tilde{C}_D} \otimes S_{\tilde{R}}})  \\
& + \frac{1}{2} vec(\mathbf{\tilde Y})^\top vec(\mathbf{S_{\tilde{R}} \tilde{Y} \mathbf{(S_{\tilde{C}_1} \otimes S_{\tilde{C}_2} \otimes \dots \otimes U_{\tilde{C}_i}^{\top} S^{-0.5}_{\Sigma_i} U^{\top}_{\Sigma_i} \Lambda^\top_i B_i \frac{\partial C_i}{\partial \theta_{\tilde{C}_i}} B_i^\top \Lambda_i U_{\Sigma_i} S^{-0.5}_{\Sigma_i} U_{\tilde{C}_i} \otimes \dots \otimes S_{\tilde{C}_D}})}),
\end{split}
\end{eqnarray*}
\normalsize
\noindent where the determinant term of the above equation is derived by computing the derivative of $\ln \left | \mathbf{K} \right |$:
\small
\begin{eqnarray*} \label{eq:determinant_c}
\begin{split}
& \frac{\partial \ln \left | \mathbf{K} \right |}{\partial \theta_{\mathbf{C}_i}} = \frac{\partial}{\partial \theta_{\mathbf{C}_i}} \ln \left | \bigotimes_{k=1}^D \mathbf{B}_k\mathbf{C}_i\mathbf{B}_k^\top \otimes \mathbf{R} + \bigotimes_{k=1}^D \mathbf{\Lambda}_k\mathbf{\Sigma}_k\mathbf{\Lambda}_k^\top \otimes \mathbf{\Omega}  \right | \\ 
& = Tr[(\bigotimes_{k=1}^D \mathbf{B}_k\mathbf{C}_k\mathbf{B}_k^\top \otimes \mathbf{R} + \bigotimes_{k=1}^D \mathbf{\Lambda}_k\mathbf{\Sigma}_k\mathbf{\Lambda}_k^\top \otimes \mathbf{\Omega})^{-1} \frac{\partial}{\partial \theta_{\mathbf{C}_i}}(\bigotimes_{k=1}^D \mathbf{B}_k\mathbf{C}_k\mathbf{B}_k^\top \otimes \mathbf{R} + \bigotimes_{k=1}^D \mathbf{\Lambda}_k\mathbf{\Sigma}_k\mathbf{\Lambda}_k^\top \otimes \mathbf{\Omega})] \\
& = Tr[(\bigotimes_{k=1}^D \mathbf{\Lambda}_k\mathbf{U_{\Sigma_k}S_{\Sigma_k}^{-0.5}} \otimes \mathbf{U_{\Omega}S_{\Omega}^{-0.5}})(\bigotimes_{k=1}^D \mathbf{\tilde{C}_k} \otimes \mathbf{\tilde{R}} + \mathbf{I})^{-1}(\bigotimes_{k=1}^D \mathbf{S_{\Sigma_k}^{-0.5}U_{\Sigma_k}^{\top}\Lambda_k^{\top}} \otimes \mathbf{S_{\Omega}^{-0.5}U_{\Omega}^{\top}}) \\ 
& (\bigotimes_{k=1}^D \mathbf{B}_k \frac{\partial \mathbf{C}}{\partial \theta_{\mathbf{C}_i}} \mathbf{B}_k^\top \otimes \mathbf{R})] \\
& = Tr[(\bigotimes_{k=1}^D \mathbf{\Lambda}_k\mathbf{U_{\Sigma_k}S_{\Sigma_k}^{-0.5}} \otimes \mathbf{U_{\Omega}S_{\Omega}^{-0.5}})(\bigotimes_{k=1}^D \mathbf{U_{\tilde C_k}} \otimes \mathbf{U_{\tilde{R}}})(\bigotimes_{k=1}^D \mathbf{S_{\tilde{C}_k}} \otimes \mathbf{S_{\tilde{R}}}+\mathbf{I})^{-1}(\bigotimes_{i=k}^D \mathbf{U_{\tilde{C}_k}^{\top}} \otimes \mathbf{U_{\tilde{R}}^{\top}}) \\ 
& (\bigotimes_{k=1}^D \mathbf{S_{\Sigma_k}^{-0.5}U_{\Sigma_k}^{\top}\Lambda_k^{\top}} \otimes \mathbf{S_{\Omega}^{-0.5}U_{\Omega}^{\top}}) (\bigotimes_{k=1}^D \mathbf{B}_k \frac{\partial \mathbf{C}}{\partial \theta_{\mathbf{C}_i}} \mathbf{B}_k^\top \otimes \mathbf{R})] \\
& = Tr[(\bigotimes_{k=1}^D \mathbf{S_{\tilde{C}_k}} \otimes \mathbf{S_{\tilde{R}}}+\mathbf{I})^{-1}(\bigotimes_{k=1}^D \mathbf{U_{\tilde{C}_k}^{\top}} \otimes \mathbf{U_{\tilde{R}}^{\top}}) 
(\bigotimes_{k=1}^D \mathbf{S_{\Sigma_k}^{-0.5}U_{\Sigma_k}^{\top}\Lambda_k^{\top}} \otimes \mathbf{S_{\Omega}^{-0.5}U_{\Omega}^{\top}}) \\
& (\bigotimes_{k=1}^D \mathbf{B}_k \frac{\partial \mathbf{C}}{\partial \theta_{\mathbf{C}_i}} \mathbf{B}_k^\top \otimes \mathbf{R})(\bigotimes_{k=1}^D \mathbf{\Lambda}_k\mathbf{U_{\Sigma_k}S_{\Sigma_k}^{-0.5}} \otimes \mathbf{U_{\Omega}S_{\Omega}^{-0.5}})(\bigotimes_{k=1}^D \mathbf{U_{\tilde C_k}} \otimes \mathbf{U_{\tilde{R}}})] \\
& = Tr[(\bigotimes_{k=1}^D \mathbf{S_{\tilde{C}_k}} \otimes \mathbf{S_{\tilde{R}}}+\mathbf{I})^{-1}(\bigotimes_{k=1}^D \mathbf{U_{\tilde{C}_k}^\top S_{\Sigma_k}^{-0.5} U_{\Sigma_k}^\top A_{k}^{\top} B_{k} \frac{\partial C_k}{\partial \theta_{C_i}} B_{k}^{\top} A_{k} U_{\Sigma_k} S_{\Sigma_k}^{-0.5} U_{\tilde{C}_k}} \\
& \otimes \mathbf{U_{\tilde{R}}^\top S_{\Omega}^{-0.5} U_{\Omega}^{\top} R U_{\Omega} S_{\Omega}^{-0.5} U_{\tilde{R}}})]
= diag((\bigotimes_{k=1}^D \mathbf{S_{\tilde{C}_k} \otimes S_{\tilde{R}} + I})^{-1})^\top \\
& diag(\mathbf{S_{\tilde{C}_1} \otimes S_{\tilde{C}_2} \otimes \dots \otimes U_{\tilde{C}_i}^{\top} S^{-0.5}_{\Sigma_i} U^{\top}_{\Sigma_i} \Lambda^\top_i B_i \frac{\partial C_i}{\partial \theta_{\tilde{C}_i}} B_i^\top \Lambda_i U_{\Sigma_i} S^{-0.5}_{\Sigma_i} U_{\tilde{C}_i} \otimes \dots \otimes S_{\tilde{C}_D} \otimes S_{\tilde{R}}}) ,
\end{split}
\end{eqnarray*}
\normalsize

\noindent and for the squared term we have:
\small
\begin{eqnarray*} \label{eq:squared_c}
\begin{split}
& \frac{\partial}{\partial \theta_{\mathbf{C}_i}} [vec(\mathbf{Y})^\top \mathbf{K}^{-1} vec(\mathbf{Y})] = vec(\mathbf{Y})^\top \frac{\partial \mathbf{K}^{-1}}{\partial \theta_{\mathbf{C}_i}} vec(\mathbf{Y}) = 
- vec(\mathbf{Y})^\top \mathbf{K}^{-1} \frac{\partial \mathbf{K}}{\partial \theta_{\mathbf{C}_i}} \mathbf{K}^{-1}vec(\mathbf{Y}) \\ 
& = - vec(\mathbf{Y})^\top (\bigotimes_{k=1}^D \mathbf{\Lambda}_k\mathbf{U_{\Sigma_k}S_{\Sigma_k}^{-0.5}} \otimes \mathbf{U_{\Omega}S_{\Omega}^{-0.5}})(\bigotimes_{k=1}^D \mathbf{U_{\tilde C_k}} \otimes \mathbf{U_{\tilde{R}}})(\bigotimes_{k=1}^D \mathbf{S_{\tilde{C}_k}} \otimes \mathbf{S_{\tilde{R}}}+\mathbf{I})^{-1}(\bigotimes_{i=k}^D \mathbf{U_{\tilde{C}_k}^{\top}} \otimes \mathbf{U_{\tilde{R}}^{\top}}) \\
& (\bigotimes_{k=1}^D \mathbf{S_{\Sigma_k}^{-0.5}U_{\Sigma_k}^{\top}\Lambda_k^{\top}} \otimes \mathbf{S_{\Omega}^{-0.5}U_{\Omega}^{\top}}) (\bigotimes_{k=1}^D \mathbf{\Lambda}_k\mathbf{U_{\Sigma_k}S_{\Sigma_k}^{0.5}} \otimes \mathbf{U_{\Omega}S_{\Omega}^{0.5}}) \\
& (\bigotimes_{k=1}^D \frac{\partial \mathbf{\tilde{C}_k }}{\partial \theta_{\mathbf{C}_i}}\otimes \mathbf{\tilde{R}})(\bigotimes_{k=1}^D \mathbf{S_{\Sigma_k}^{0.5}U_{\Sigma_k}^{\top}\Lambda_k^{\top}} \otimes \mathbf{S_{\Omega}^{0.5}U_{\Omega}^{\top}})(\bigotimes_{k=1}^D \mathbf{\Lambda}_k\mathbf{U_{\Sigma_k}S_{\Sigma_k}^{-0.5}} \otimes \mathbf{U_{\Omega}S_{\Omega}^{-0.5}}) \\
&(\bigotimes_{k=1}^D \mathbf{U_{\tilde C_k}} \otimes \mathbf{U_{\tilde{R}}})(\bigotimes_{k=1}^D \mathbf{S_{\tilde{C}_k}} \otimes \mathbf{S_{\tilde{R}}}+\mathbf{I})^{-1}(\bigotimes_{i=k}^D \mathbf{U_{\tilde{C}_k}^{\top}} \otimes \mathbf{U_{\tilde{R}}^{\top}})(\bigotimes_{k=1}^D \mathbf{S_{\Sigma_k}^{-0.5}U_{\Sigma_k}^{\top}\Lambda_k^{\top}} \otimes \mathbf{S_{\Omega}^{-0.5}U_{\Omega}^{\top}}) vec(\mathbf{Y}) \\
& = -vec(\mathbf{\tilde Y})^\top (\bigotimes_{k=1}^D \mathbf{U_{\tilde{C}_k}^\top} \otimes \mathbf{U_{\tilde{R}}^\top}) (\bigotimes_{k=1}^D \frac{\partial \mathbf{\tilde{C}_k }}{\partial \theta_{\mathbf{C}_i}}\otimes \mathbf{\tilde{R}}) (\bigotimes_{k=1}^D \mathbf{U_{\tilde{C}_k}} \otimes \mathbf{U_{\tilde{R}}}) vec(\mathbf{\tilde Y}) \\
& = -vec(\mathbf{\tilde Y})^\top (\bigotimes_{k=1}^D \mathbf{U_{\tilde{C}_k}^\top S_{\Sigma_k}^{-0.5} U_{\Sigma_k}^\top A_{k}^{\top} B_{k} \frac{\partial C_k}{\partial \theta_{C_i}} B_{k}^{\top} A_{k} U_{\Sigma_k} S_{\Sigma_k}^{-0.5} U_{\tilde{C}_k} \otimes  S_{\tilde R}}) vec(\mathbf{\tilde Y}) \\
& = -vec(\mathbf{\tilde Y})^\top (\mathbf{S_{\tilde R} \tilde{Y} \bigotimes_{k=1}^D U_{\tilde{C}_k}^\top S_{\Sigma_k}^{-0.5} U_{\Sigma_k}^\top A_{k}^{\top} B_{k} \frac{\partial C_k}{\partial \theta_{C_i}} B_{k}^{\top} A_{k} U_{\Sigma_k} S_{\Sigma_k}^{-0.5} U_{\tilde{C}_k}}) \\
& -vec(\mathbf{\tilde Y})^\top vec(\mathbf{S_{\tilde{R}} \tilde{Y} \mathbf{(S_{\tilde{C}_1} \otimes S_{\tilde{C}_2} \otimes \dots \otimes U_{\tilde{C}_i}^{\top} S^{-0.5}_{\Sigma_i} U^{\top}_{\Sigma_i} \Lambda^\top_i B_i \frac{\partial C_i}{\partial \theta_{\tilde{C}_i}} B_i^\top \Lambda_i U_{\Sigma_i} S^{-0.5}_{\Sigma_i} U_{\tilde{C}_i} \otimes \dots \otimes S_{\tilde{C}_D}})}).
\end{split}
\end{eqnarray*}
\normalsize

\subsubsection*{Gradients of $L$ with Respect to $\theta_{\mathbf{\Sigma}_i}$}
\label{subsubsec:gradient_s}
\scriptsize
\begin{eqnarray*} \label{eq:derivatives_s}
\begin{split}
& \frac{\partial L}{\partial \theta_{\mathbf{\Sigma}_i}}  = -\frac{1}{2}diag((\bigotimes_{k=1}^D \mathbf{S_{\tilde{\Sigma}_k} \otimes S_{\tilde{\Omega}} + I})^{-1})^\top \\
& diag(\mathbf{S_{\tilde{\Sigma}_1} \otimes S_{\tilde{\Sigma}_2} \otimes \dots \otimes U_{\tilde{\Sigma}_i}^{\top} S^{-0.5}_{C_i} U^{\top}_{C_i} B^\top_i \Lambda_i \frac{\partial \Sigma_i}{\partial \theta_{\tilde{\Sigma}_i}} \Lambda_i^\top B_i U_{C_i} S^{-0.5}_{C_i} U_{\tilde{\Sigma}_i} \otimes \dots \otimes S_{\tilde{\Sigma}_D} \otimes S_{\tilde{\Omega}}})  \\
& + \frac{1}{2} vec(\mathbf{\tilde Y})^\top vec(\mathbf{S_{\tilde{\Omega}} \tilde{Y} \mathbf{(S_{\tilde{\Sigma}_1} \otimes S_{\tilde{\Sigma}_2} \otimes \dots \otimes U_{\tilde{\Sigma}_i}^{\top} S^{-0.5}_{C_i} U^{\top}_{C_i} B^\top_i \Lambda_i \frac{\partial \Sigma_i}{\partial \theta_{\tilde{\Sigma}_i}} \Lambda_i^\top B_i U_{C_i} S^{-0.5}_{C_i} U_{\tilde{\Sigma}_i} \otimes \dots \otimes S_{\tilde{\Sigma}_D}})}),
\end{split}
\end{eqnarray*}
\normalsize
The derivation of the determinant and squared terms of $\frac{\partial L}{\partial \theta_{\mathbf{\Sigma}_i}}$ are similar to those of  $\frac{\partial L}{\partial \theta_{\mathbf{C}_i}}$.

\subsubsection*{Gradients of $L$ with Respect to $\theta_\mathbf{R}$:}
\label{subsubsec:gradient_c}
\scriptsize
\begin{eqnarray*} \label{eq:derivatives_r}
\begin{split}
\frac{\partial L}{\partial \theta_\mathbf{R}} = & -\frac{1}{2}diag((\bigotimes_{k=1}^D \mathbf{S_{\tilde{C}_k} \otimes S_{\tilde{R}} + I})^{-1})^\top diag(\bigotimes_{k=1}^D \mathbf{S_{\tilde{C}_k}} \otimes \mathbf{U_{\tilde{R}}^\top S_{\Omega}^{-0.5} U_{\Omega}^\top}\frac{\partial \mathbf{R}}{\partial \theta_\mathbf{R}}\mathbf{U_{\Omega} S_{\Omega}^{-0.5}U_{\tilde{R}}}) \\
& + \frac{1}{2} vec(\mathbf{\tilde Y})^\top vec(\mathbf{U_{\tilde{R}}^\top S_{\Omega}^{-0.5} U_{\Omega}^\top}\frac{\partial \mathbf{R}}{\partial \theta_\mathbf{R}}\mathbf{U_{\Omega} S_{\Omega}^{-0.5}U_{\tilde{R}} \tilde Y } \bigotimes_{k=1}^D \mathbf{S_{\tilde{C}_k}}).
\end{split}
\end{eqnarray*}
\normalsize
The derivation of the determinant and squared terms of $\frac{\partial L}{\partial \theta_{\mathbf{R}}}$ are similar to those of  $\frac{\partial L}{\partial \theta_{\mathbf{C}_i}}$.

\subsubsection*{Gradients of $L$ with Respect to $\theta_\mathbf{\Omega}$:}
\label{subsubsec:gradient_c}
\scriptsize
\begin{eqnarray*} \label{eq:derivatives_r}
\begin{split}
\frac{\partial L}{\partial \theta_\mathbf{\Omega}} = & -\frac{1}{2}diag((\bigotimes_{k=1}^D \mathbf{S_{\tilde{\Sigma}_k} \otimes S_{\tilde{\Omega}} + I})^{-1})^\top diag(\bigotimes_{k=1}^D \mathbf{S_{\tilde{\Sigma}_k}} \otimes \mathbf{U_{\tilde{\Omega}}^\top S_{R}^{-0.5} U_{R}^\top}\frac{\partial \mathbf{\Omega}}{\partial \theta_\mathbf{\Omega}}\mathbf{U_{R} S_{R}^{-0.5} U_{\tilde{\Omega}}}) \\
& + \frac{1}{2} vec(\mathbf{\tilde Y})^\top vec(\mathbf{U_{\tilde{\Omega}}^\top S_{R}^{-0.5} U_{R}^\top}\frac{\partial \mathbf{\Omega}}{\partial \theta_\mathbf{\Omega}}\mathbf{U_{R} S_{R}^{-0.5}U_{\tilde{\Omega}} \tilde Y} \bigotimes_{k=1}^D \mathbf{S_{\tilde{\Sigma}_k}}).
\end{split}
\end{eqnarray*}
\normalsize
The procedure to derive the determinant and squared terms of $\frac{\partial L}{\partial \theta_{\mathbf{\Omega}}}$ is similar to $\frac{\partial L}{\partial \theta_{\mathbf{C}_i}}$.

\subsection*{Comparing the Regression Performance}
This figure summarizes the average regression performance ($R^2$) across all voxels for benchmarked approaches. All methods show similar performance in terms of the quality of regression. Note that the low  $R^2$ values are due to averaging over all voxels that many are irrelevant to regressors in $\mathbf{X}$. 
\label{subsec:regression_results}
\begin{figure*}[h]
	\centering
	\includegraphics[width=0.8\textwidth]{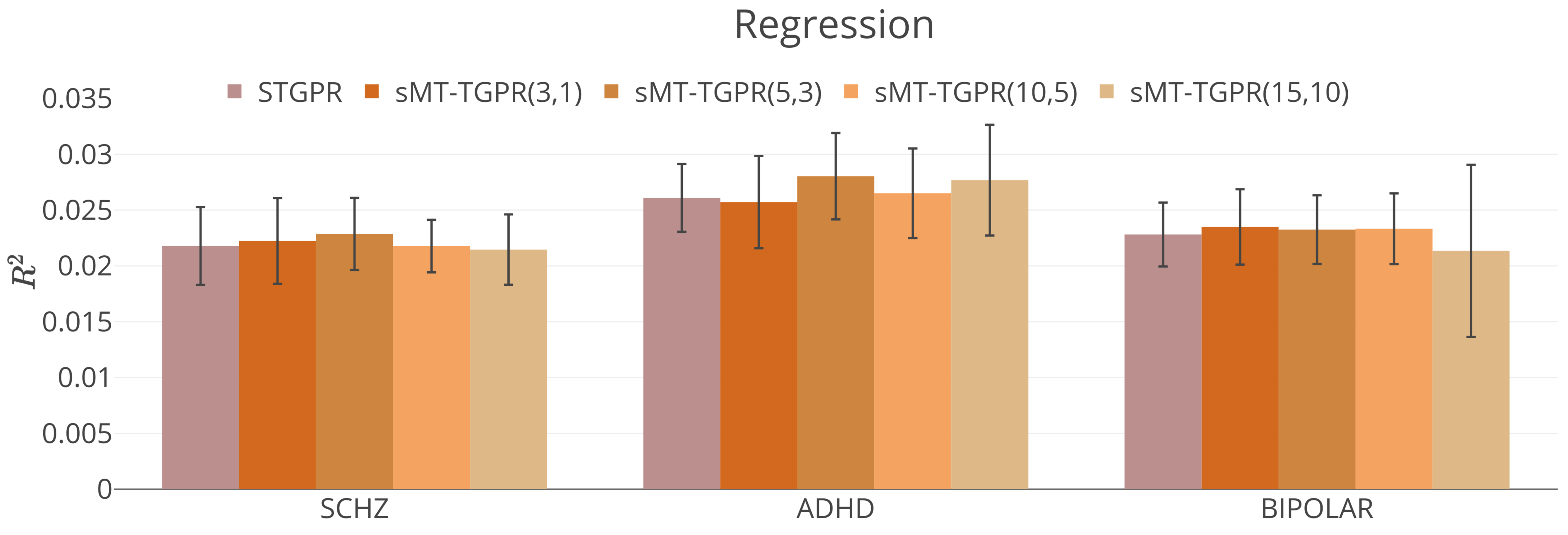}
	\caption{Comparison between ST-GPR and sMT-GPTR in terms of their regression performance.}
	\label{fig:r2}
\end{figure*}

\subsection*{Supplementary Deviation Maps}
The following figure presents a complementary results for Sec.~\ref{subsec:results_brainplots} of the main text. 
\label{subsec:brain_plots_suppl}
\begin{figure*}[h]
	\centering
	\includegraphics[width=0.75\textwidth]{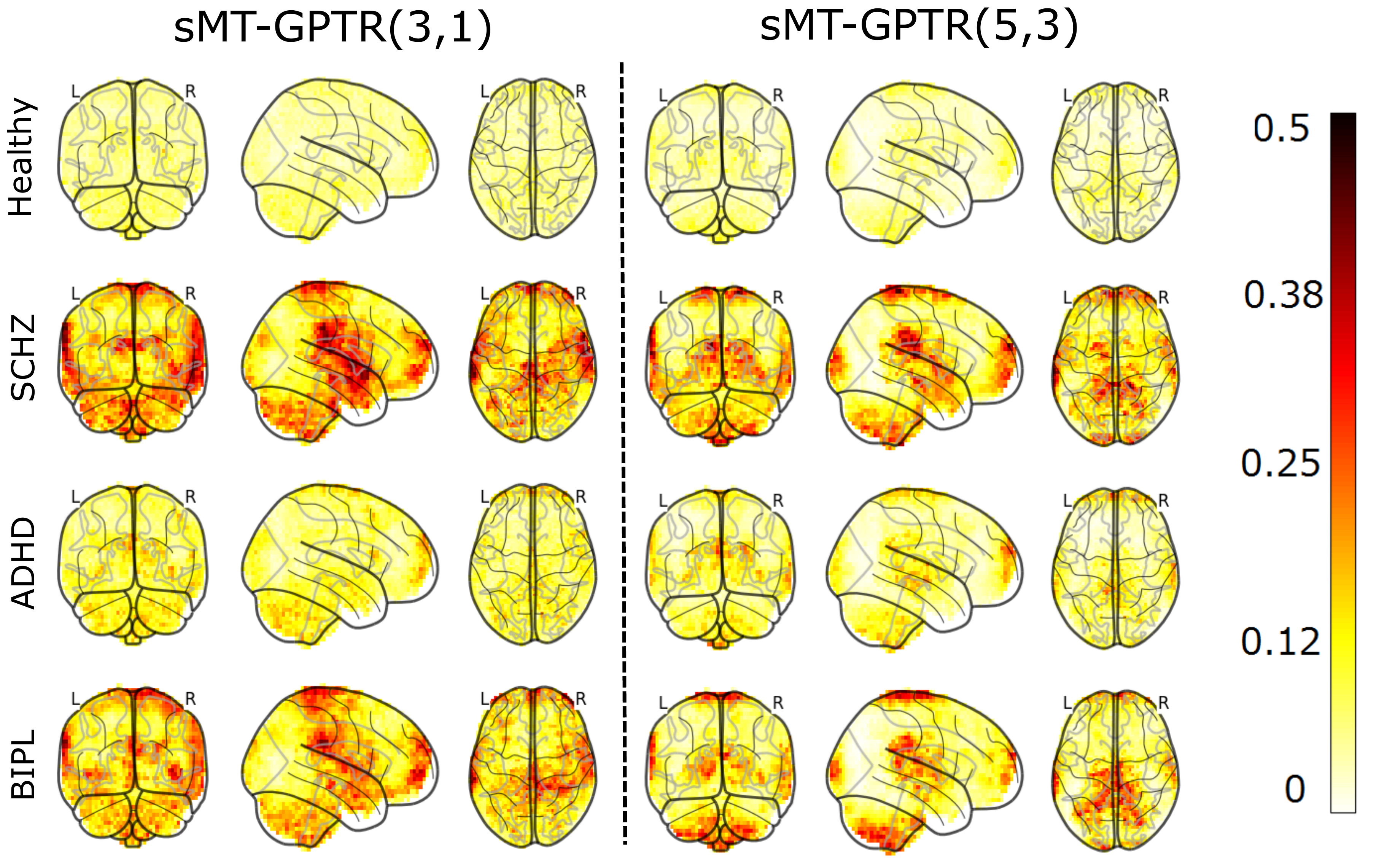}
	\caption{The probability of each voxel to deviate from the normative model in healthy and patient populations; derived by sMT-GPTR(3,1) and sMT-GPTR(5,3).}
	\label{fig:r2}
\end{figure*}
\end{document}